\documentclass[10pt,twocolumn,letterpaper]{article}

\usepackage[pagenumbers]{wacv} 

\usepackage{graphicx}
\usepackage{amsmath}
\usepackage{amssymb}
\usepackage{booktabs}
\usepackage{xr-hyper} 

\usepackage[pagebackref,breaklinks,colorlinks]{hyperref}
\usepackage{mathtools}
\usepackage{amsmath}
\usepackage{bm}
\usepackage{tabularx} 
\usepackage{multirow}
\usepackage{tikz}
\usetikzlibrary{tikzmark}
\usepackage[
  separate-uncertainty = true,
  multi-part-units = repeat
]{siunitx}
\usepackage{acronym}
\usepackage{colortbl}
\usepackage{adjustbox}

\newacro{DGM}{deep generative model}
\newacro{GAN}{generative adversarial network}
\newacro{ADM}{ablated diffusion model}
\newacro{DM}{diffusion model}
\newacro{VAE}{variational autoencoder}
\newacro{SOTA}{state-of-the-art}
\newacro{ours}[STAY Diffusion]{\textbf{ST}yled L\textbf{AY}out Diffusion}
\newacro{ean}[EA Norm]{Edge-Aware Normalization}
\newacro{sma}[SM Attention]{Styled-Mask Attention}
\newacro{VG}{Visual Genome}
\newacro{LTGM}{large text-to-image generation model}
\newacro{L2I}{layout-to-image}
\newacro{LayoutDM}{LayoutDiffusion}

\newcommand{\figteaser}{
\begin{figure*}[!tbp]
\centering
\includegraphics[width=0.8\linewidth]{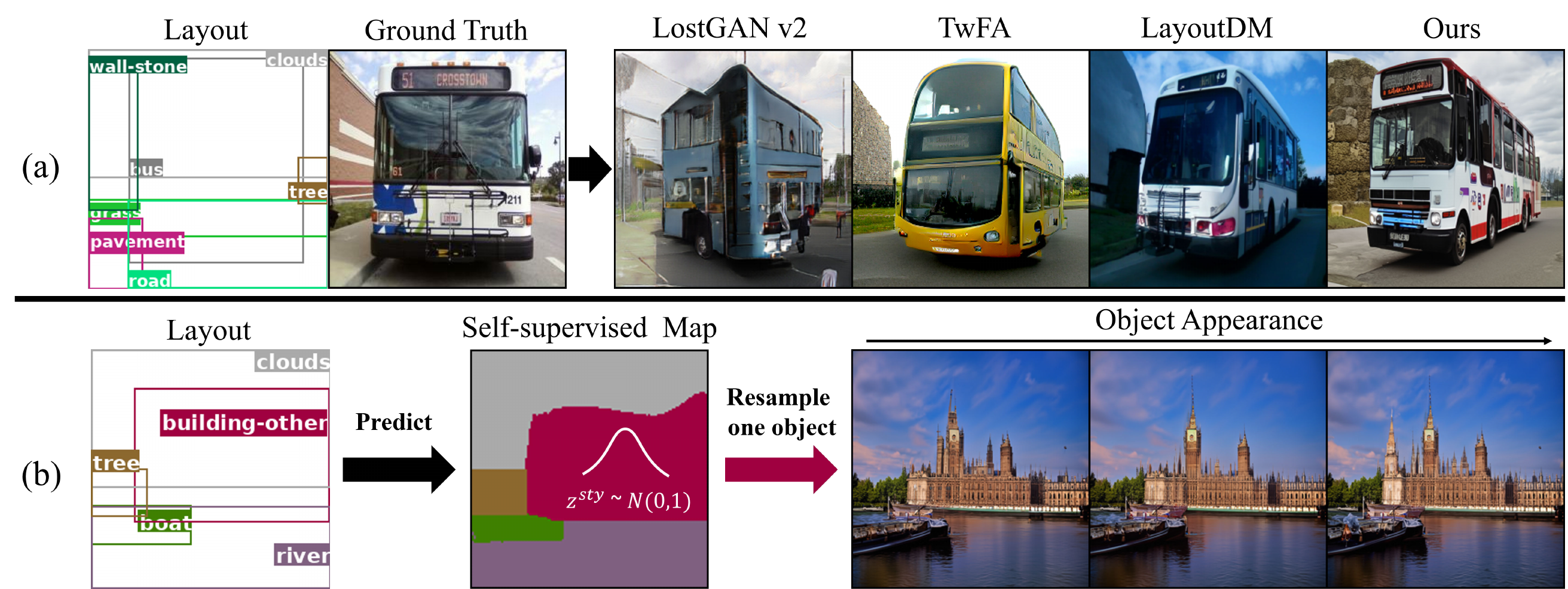}\\
\vspace{-0.2cm}
\caption{\ac{ours} can produce high-quality images based on given bounding box layouts. (a) A comparison between \ac{ours} and previous \acl{SOTA} methods. (b) In addition, \ac{ours} learns a self-supervised semantic map for each layout and can manipulate object appearance by resampling its associated latent code $z^{sty}$ (cf. the towers of the building).}
\label{fig:teaser}
\end{figure*}
}

\newcommand{\figmodel}{
\begin{figure*}[!tbp]
\centering
\includegraphics[width=1.0\linewidth]{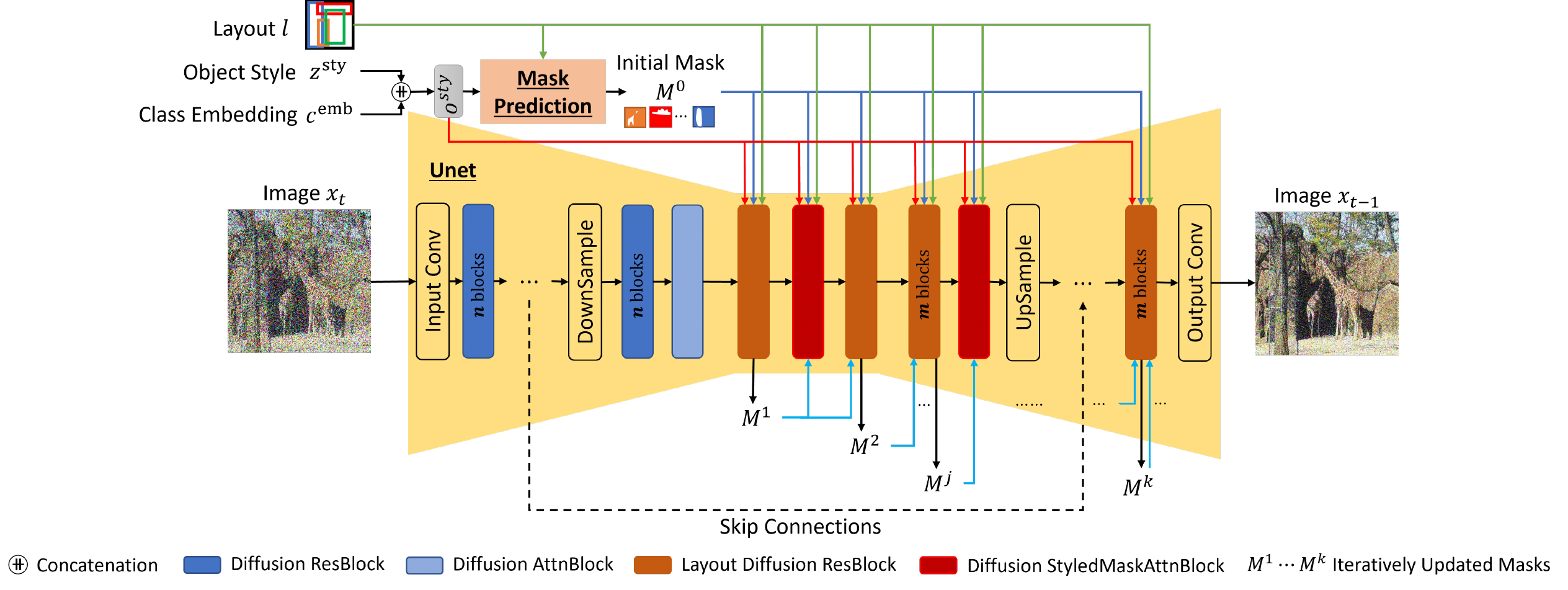}\\
\vspace{-0.4cm}
\caption{The architecture of the proposed \ac{ours}. Through the guidance of the given layout, the learnable object representations $o^{sty}$, and their respective initial masks $M^0$, the model gradually turns a noisy image into a realistic real-world scene.} 
\label{fig:model}
\end{figure*}
}

\newcommand{\figeanorm}{
\begin{figure*}[!tbp]
\centering
\includegraphics[width=0.9\linewidth]{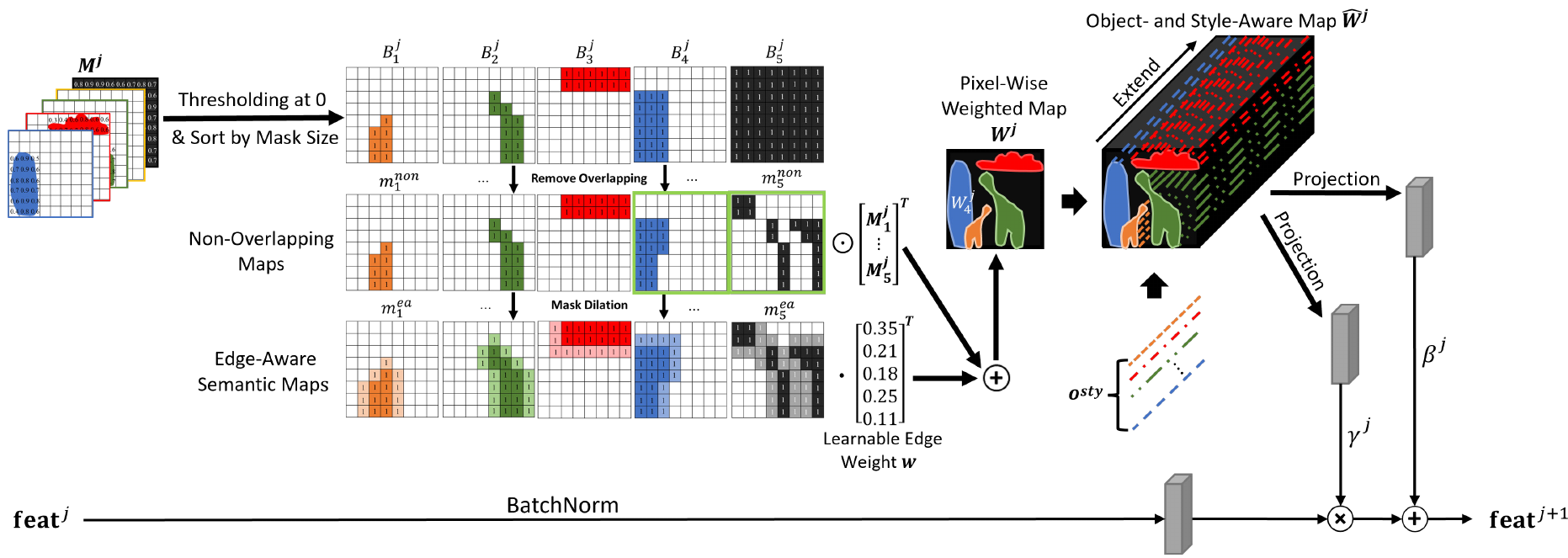}\\
\vspace{-0.4cm}
\caption{The workflow of the \acl{ean}, where object masks $M^j$ are carefully assembled into a pixel-wise weighted map $W^j$ and then extended by $o^\mathrm{sty}$ for computing the modulation parameters $\gamma$ and $\beta$. Note that in $m^\mathrm{non}_4$ and $m^\mathrm{non}_5$ (highlighted in green), some pixels are removed (i.e., set to 0) due to overlapping with other smaller objects. See text for more details.}
\label{fig:eanorm}
\end{figure*}
}

\newcommand{\figqual}{
\begin{figure*}[!tbp]
\centering
\includegraphics[width=1.0\linewidth]{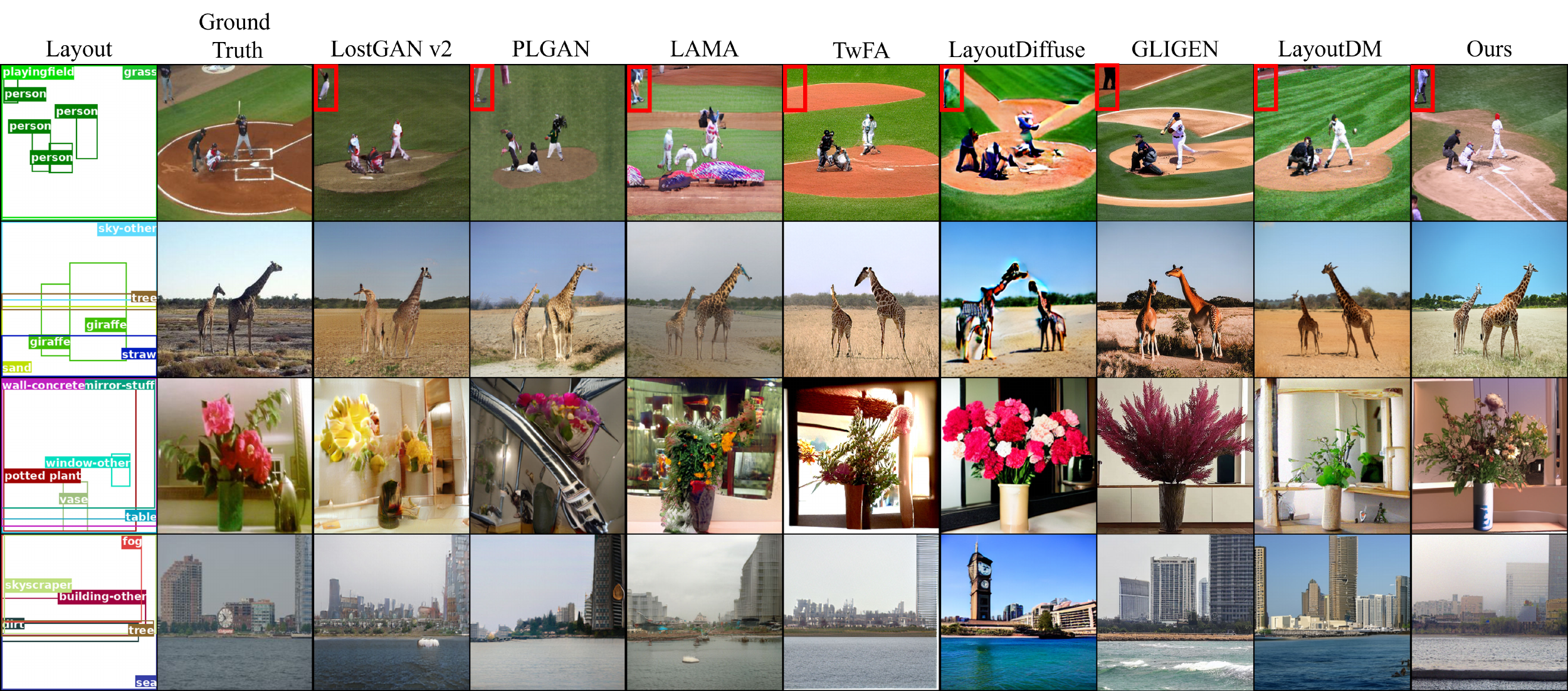}\\
\vspace{-0.3cm}
\caption{Qualitative comparison to the \ac{SOTA} methods on COCO-stuff 256 $\times$ 256. \ac{ours} shows better controllability and object recognizability over previous methods (e.g., the people on the playfield in the first row and the foggy effect in the last row). Zoom in for better view.}
\label{fig:qual}
\vspace{-0.7cm}
\end{figure*}
}

\newcommand{\figsty}{
\begin{figure*}[!tbp]
    \centering
    \begin{minipage}{0.49\textwidth}
        \centering
        \includegraphics[width=0.9\textwidth]{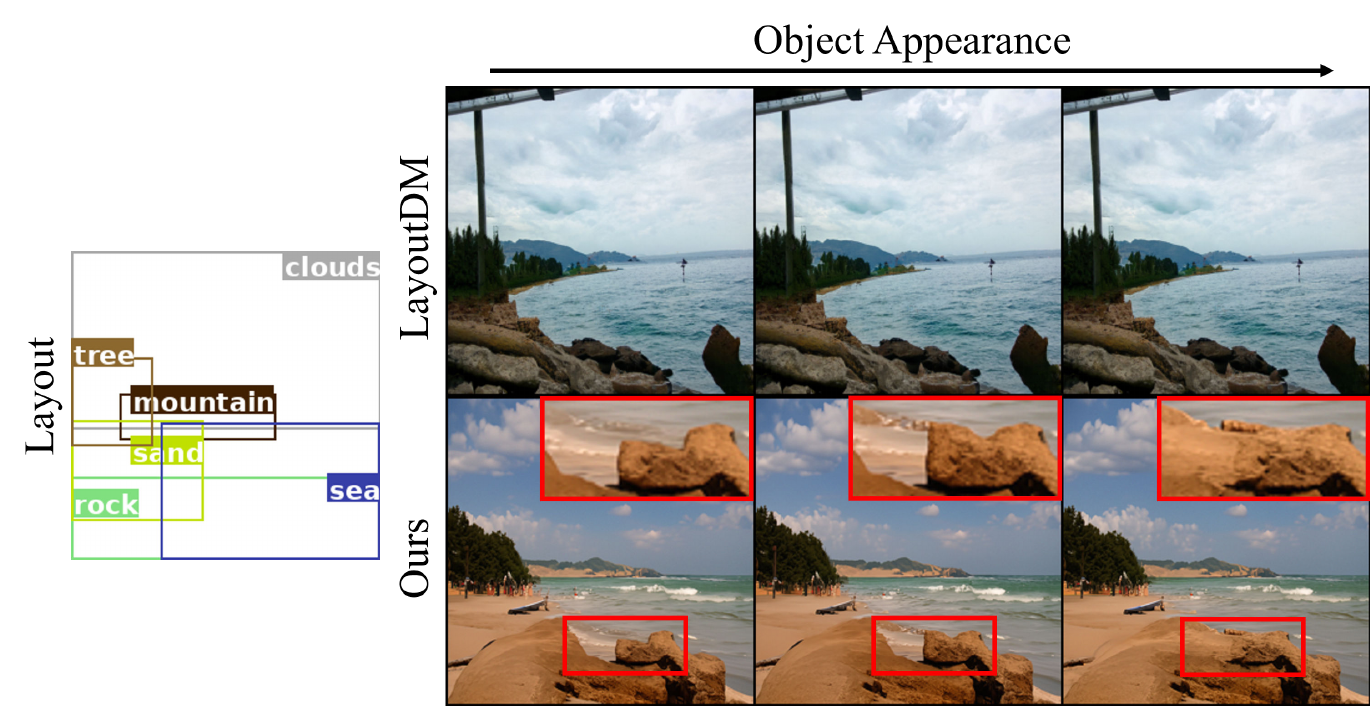} 
        \vspace{-0.3cm}
        \caption{Given the same layout and input noise, \ac{ours} can manipulate the appearance of an object (cf. the rock) by resampling its associated $z^\mathrm{sty}$ (Zoom in for better view).}
        \label{fig:stya}
    \end{minipage}\hfill
    \vspace{-0.2cm}
    \begin{minipage}{0.49\textwidth}
        \centering
        \includegraphics[width=0.9\textwidth]{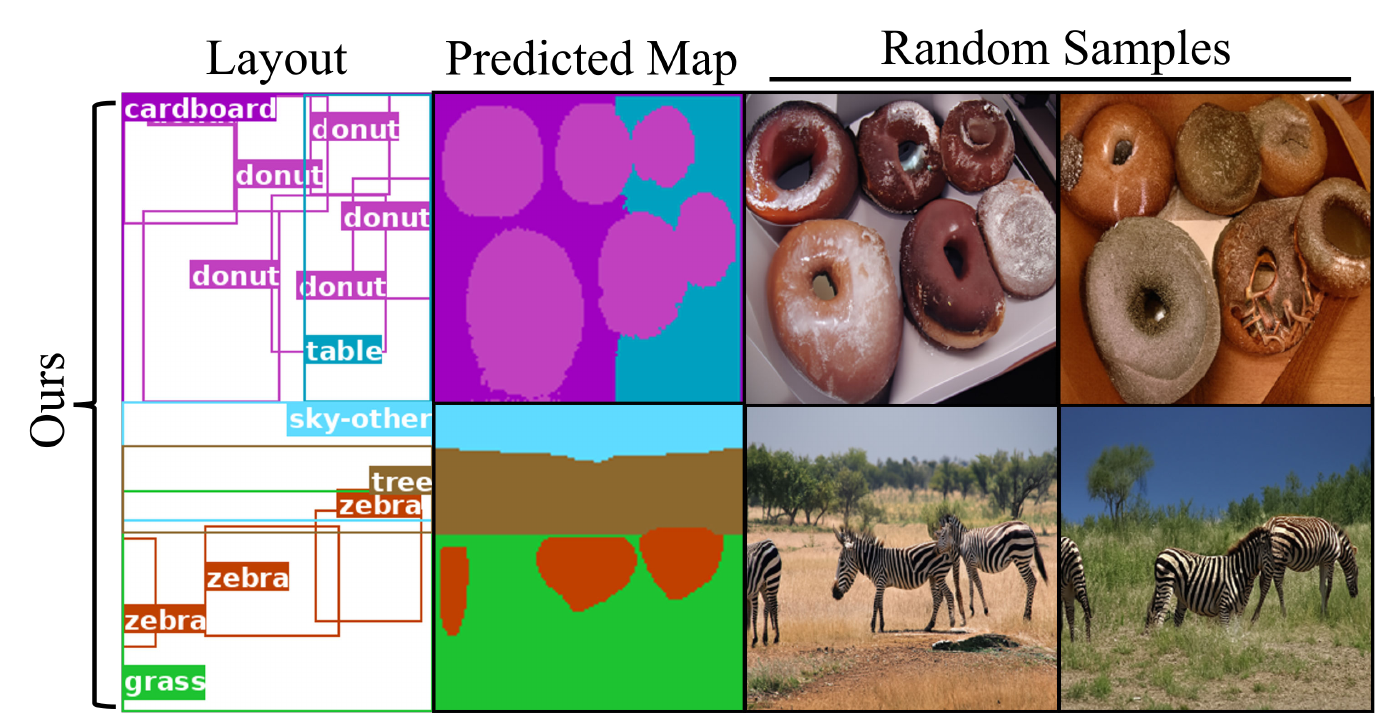} 
        \vspace{-0.3cm}
        \caption{The self-supervised maps and their generated images. We randomly sampled two input noises to show the output diversity.}
        \label{fig:maskb}
    \end{minipage}
\vspace{-0.2cm}
\end{figure*}
}

\newcommand{\figablation}{
\begin{figure*}[!htbp]
\centering
\includegraphics[width=1.0\linewidth]{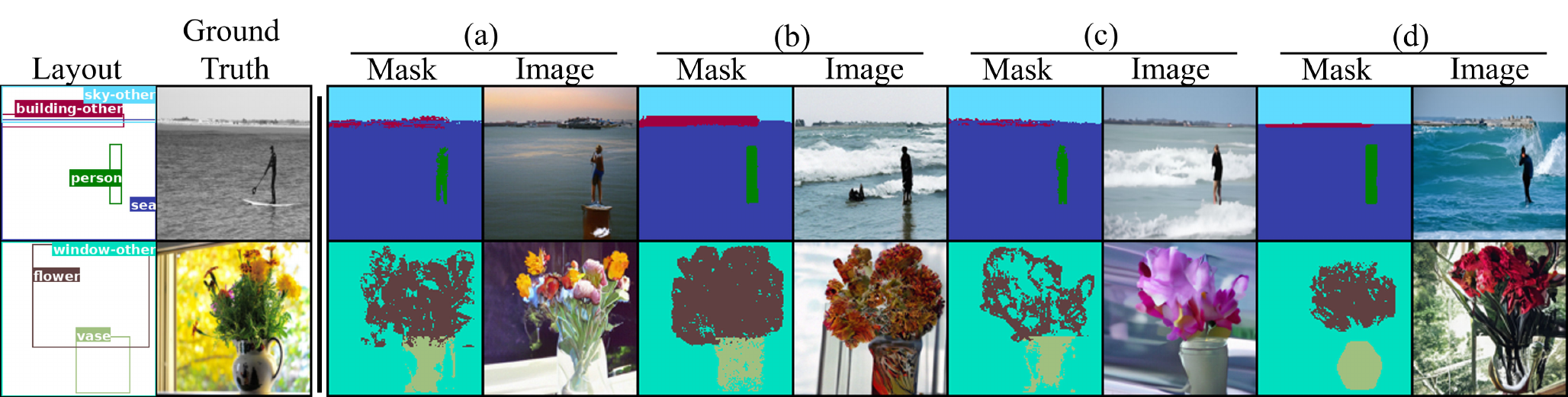}\\
\vspace{-0.3cm}
\caption{The generated images and their predicted self-supervised semantic maps from the ablated models in Tab.~\ref{tab:ablation}.}
\label{fig:ablation}
\end{figure*}
}

\newcommand{\figblockss}{
\begin{figure}[!htbp]
\begin{subfigure}[b]{0.5\textwidth}
\centering
\includegraphics[width=0.9\textwidth]{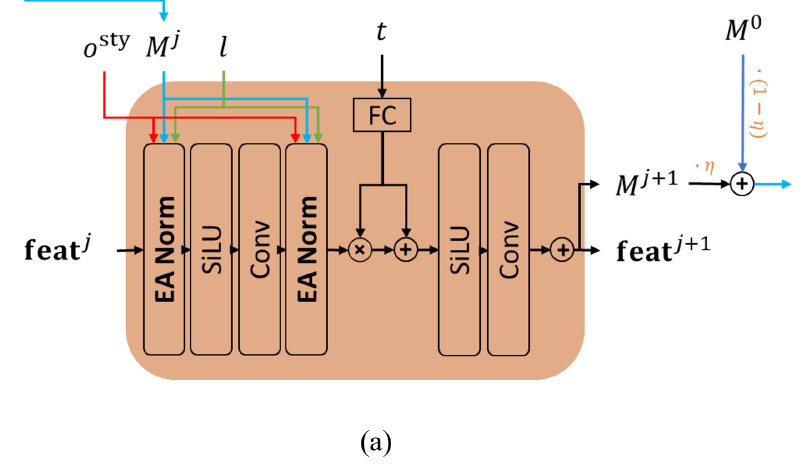}
\vspace{0.3cm}
\end{subfigure}
\begin{subfigure}[b]{0.5\textwidth}
\centering
\includegraphics[width=0.8\textwidth]{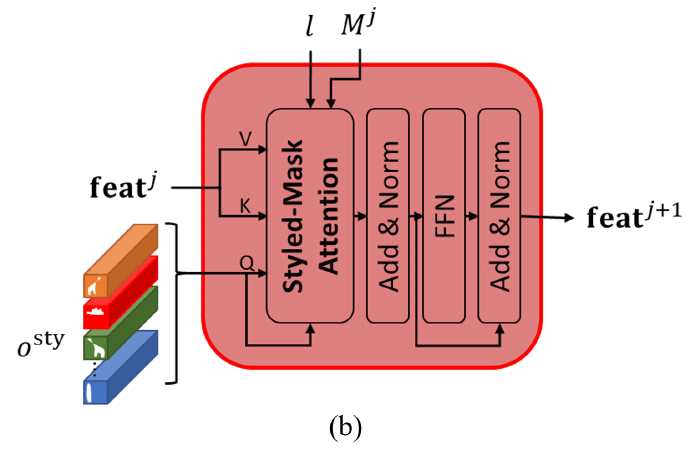}
\end{subfigure}
\vspace{-0.7cm}
\caption{The design of (a) the Layout Diffusion Resblock and (b) the Diffusion StyledMaskAttnBlock, respectively.}
\label{fig:blockss}
\end{figure}
}

\newcommand{\figcombadd}{
\begin{figure*}[!tbp]
\centering
\includegraphics[width=1.0\linewidth]{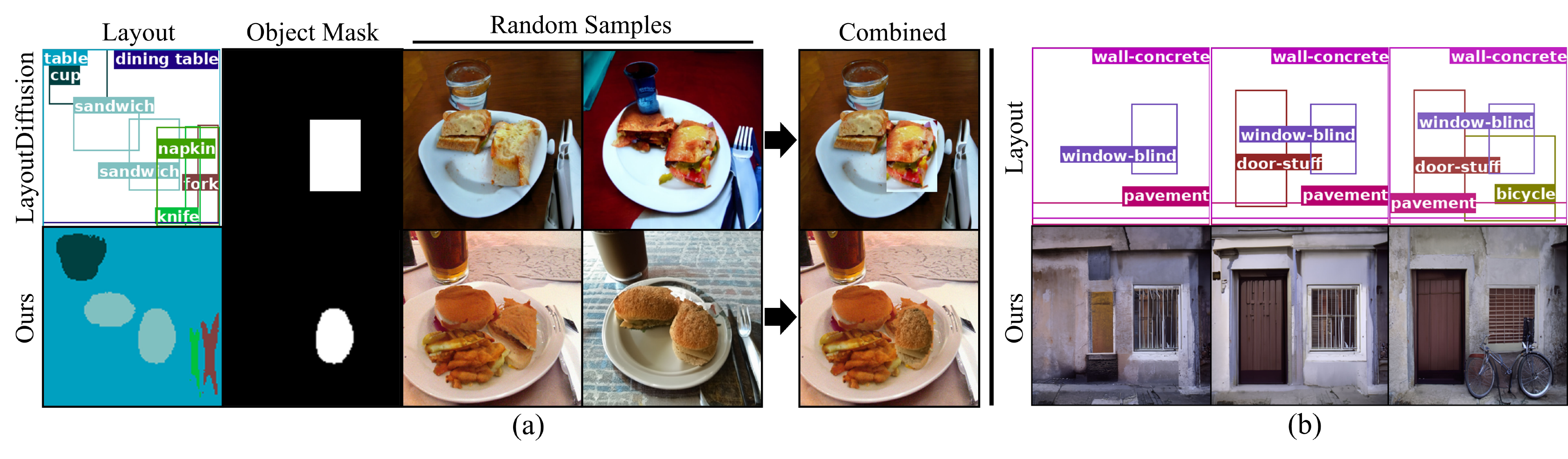}\\
\vspace{-0.5cm}
\caption{The interactivity of \ac{ours}. (a) With the self-supervised semantic maps, \ac{ours} provides more accurate object location for tasks like image blending.  (b) \ac{ours} can adapt to reconfigured layouts. }
\label{fig:combadd}
\end{figure*}
}

\newcommand{\figqualtwo}{
\begin{figure*}[!tbp]
\centering
\includegraphics[width=0.8\linewidth]{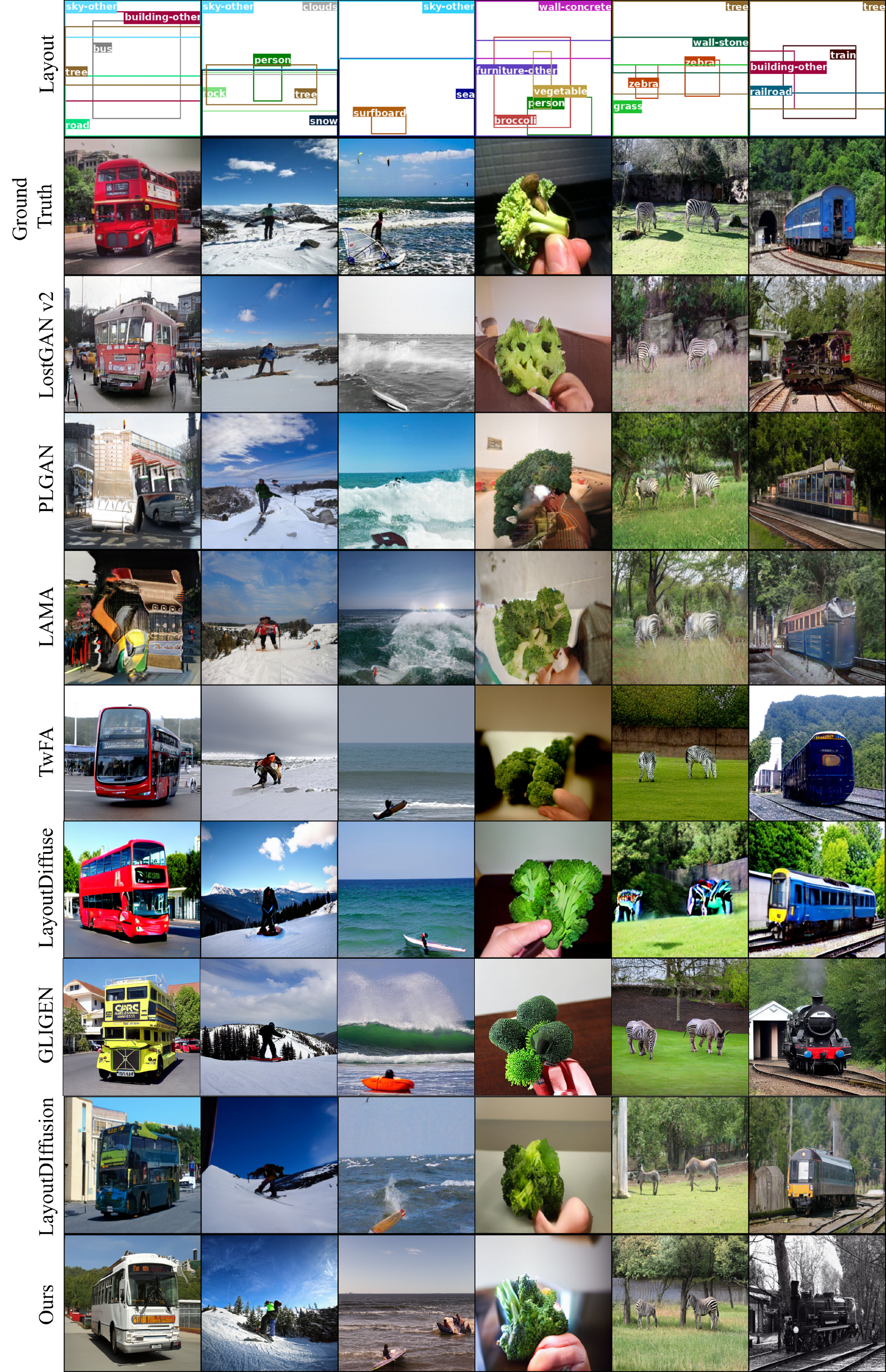}\\
\caption{More comparison with previous methods on COCO-stuff 256 $\times$ 256. Zoom in for better view.}
\label{fig:qualtwo}
\end{figure*}
}
\newcommand{\figqualthree}{
\begin{figure*}[!tbp]
\centering
\includegraphics[width=0.8\linewidth]{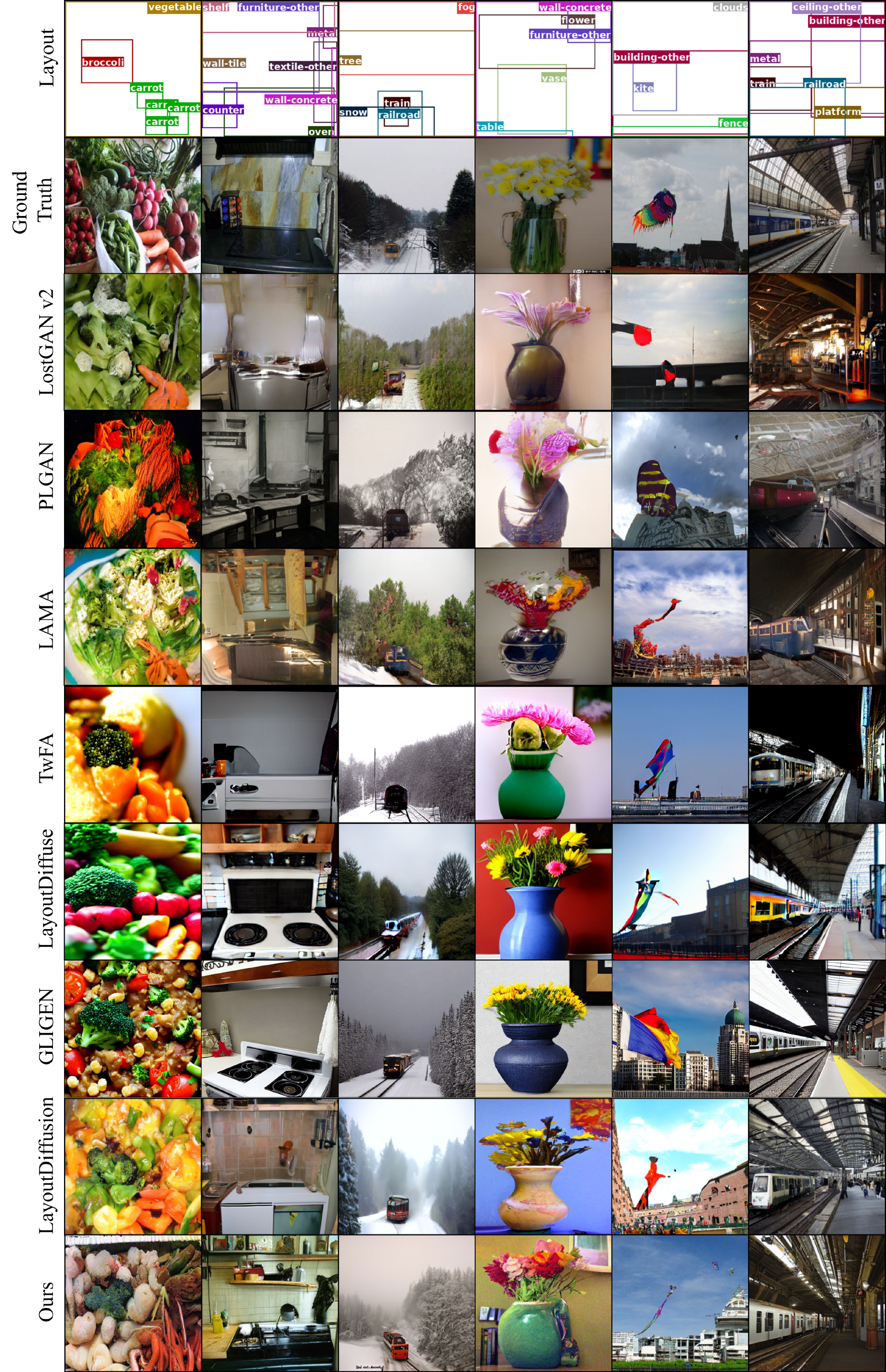}\\
\caption{More comparison with previous methods on COCO-stuff 256 $\times$ 256. Zoom in for better view.}
\label{fig:qualthree}
\end{figure*}
}

\newcommand{\figvgqualone}{
\begin{figure*}[!tbp]
\centering
\includegraphics[width=0.9\linewidth]{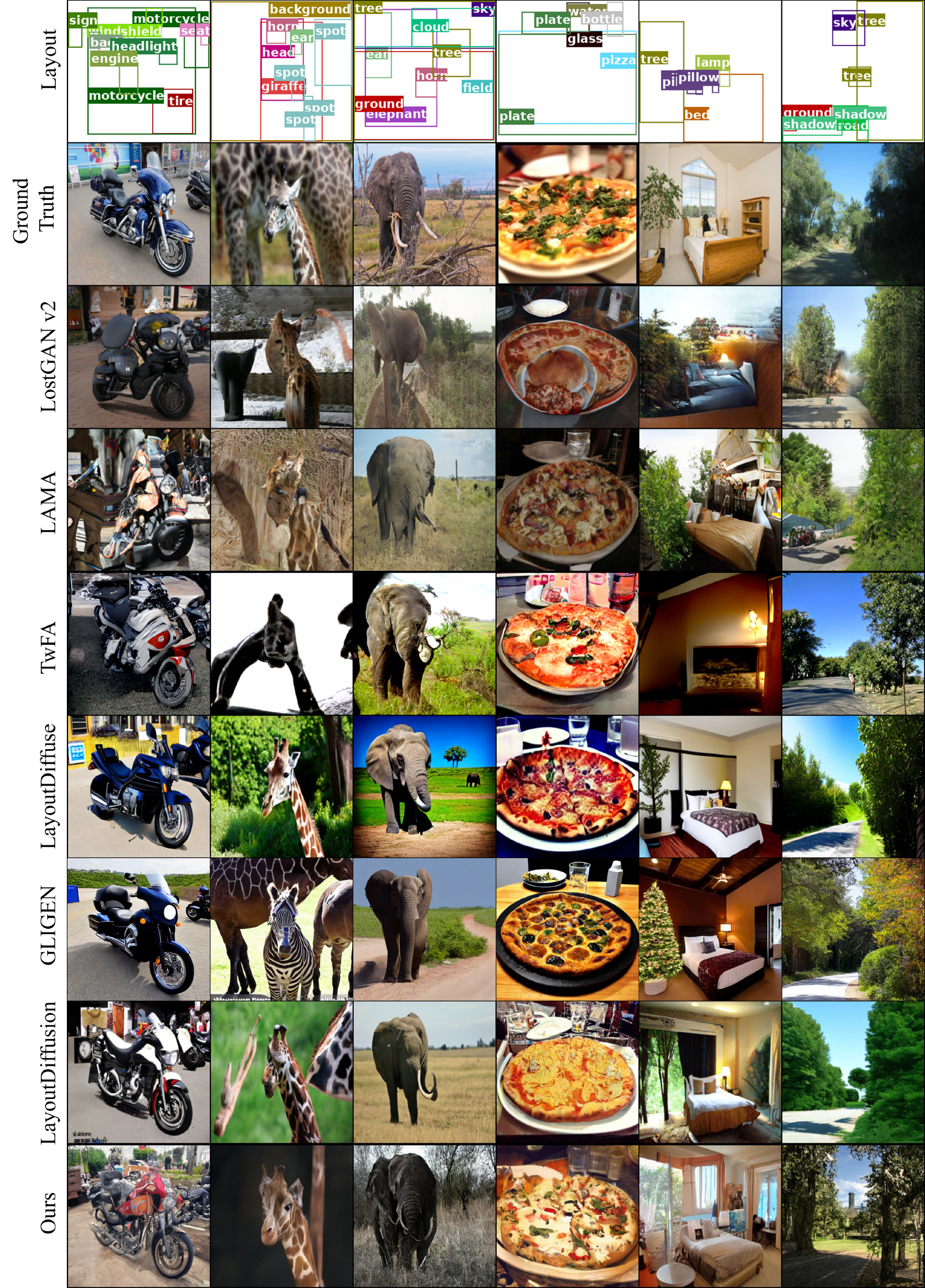}\\
\caption{Visual comparison with previous methods on \ac{VG} 256 $\times$ 256.}
\label{fig:vgqualone}
\end{figure*}
}
\newcommand{\figvgqualtwo}{
\begin{figure*}[!tbp]
\centering
\includegraphics[width=0.9\linewidth]{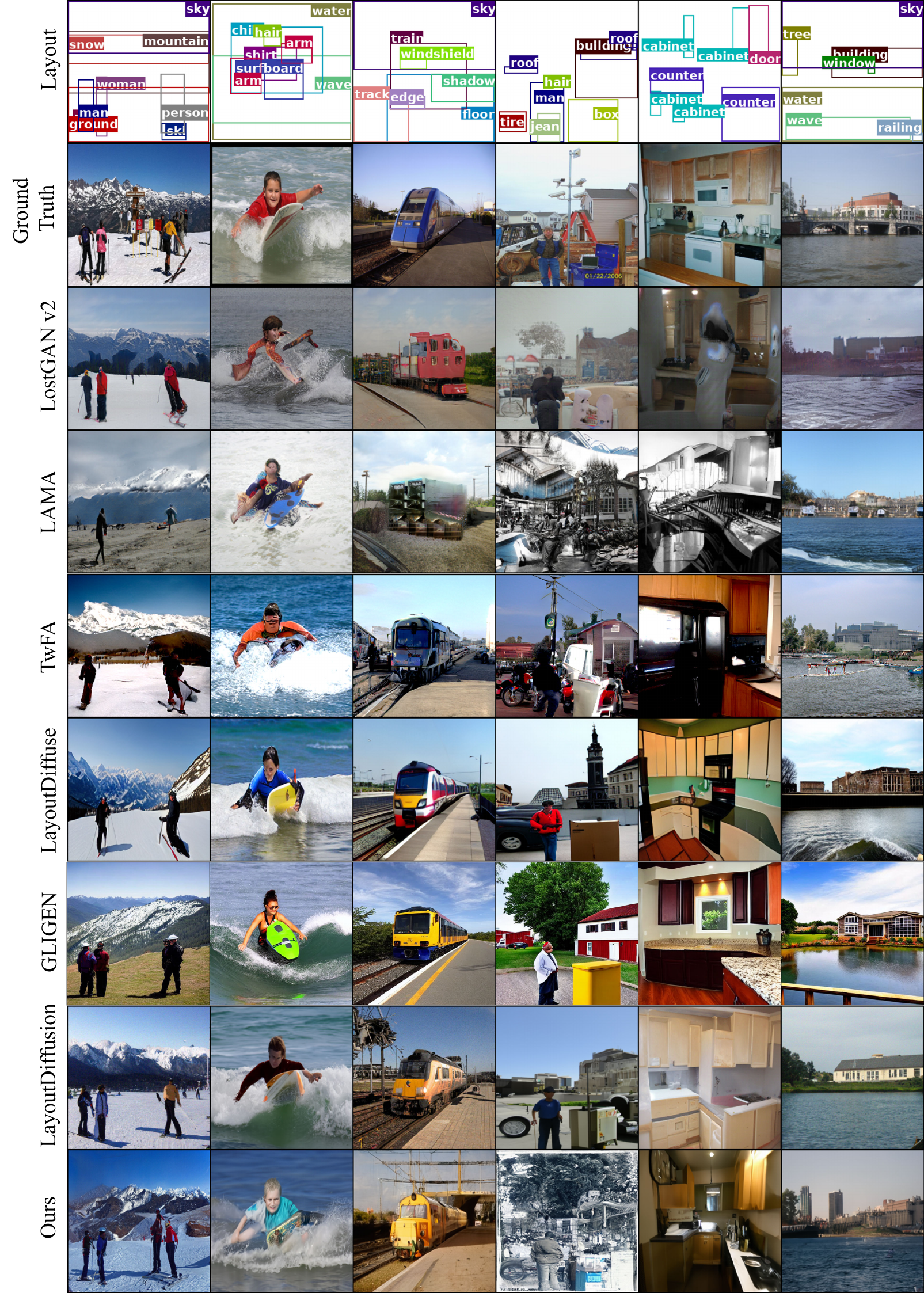}\\
\caption{Visual comparison with previous methods on \ac{VG} 256 $\times$ 256.}
\label{fig:vgqualtwo}
\end{figure*}
}

\newcommand{\figstytwo}{
\begin{figure*}[!tbp]
\centering
\includegraphics[width=0.7\linewidth]{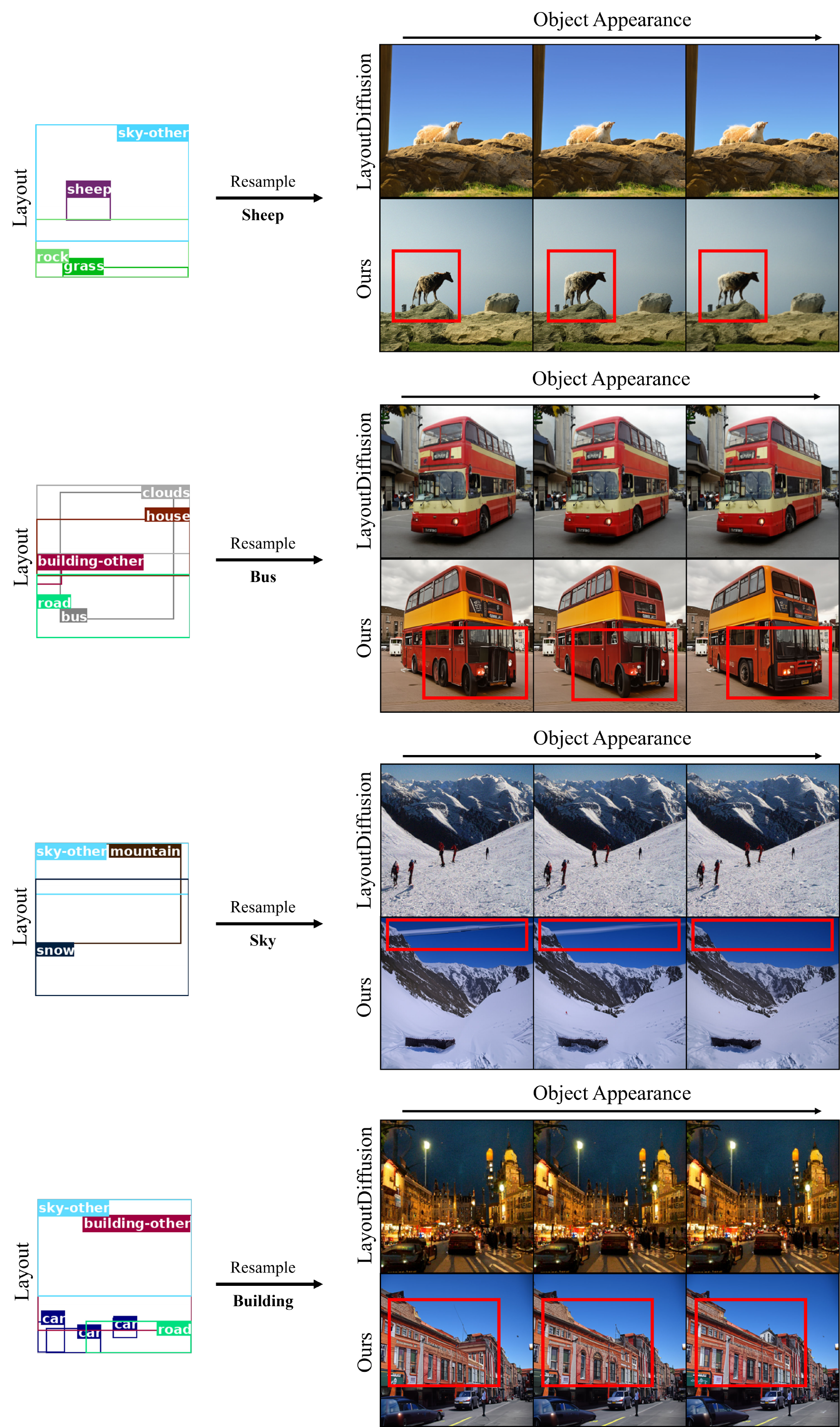}\\
\caption{The demonstration of fine-grained style variations offered by our \ac{ours}. Note that only one object is resampled in each image (Zoom in for better view).}
\label{fig:stytwo}
\end{figure*}
}

\newcommand{\figmasks}{
\begin{figure*}[!tbp]
\centering
\includegraphics[width=0.9\linewidth]{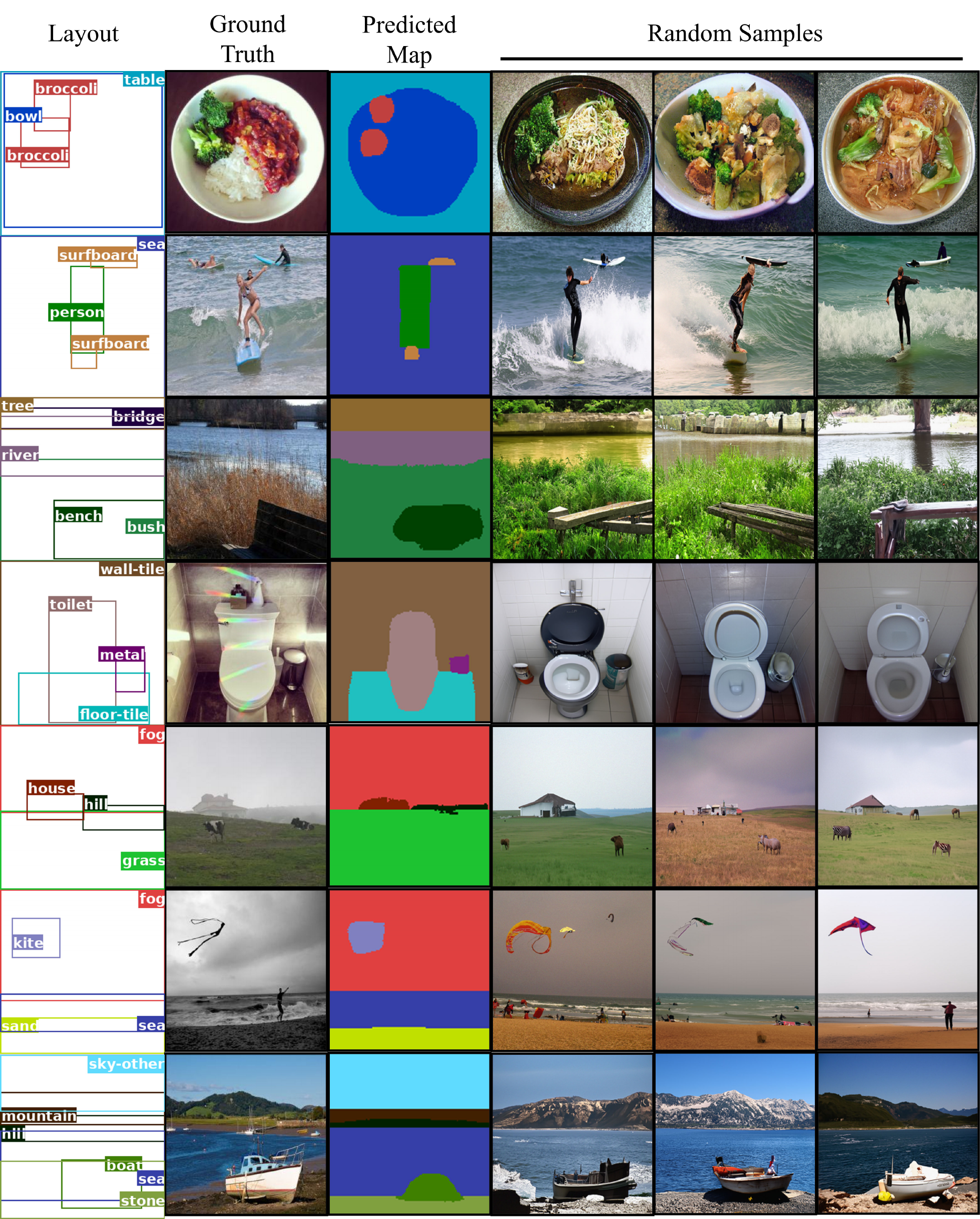}\\
\caption{The demonstration of self-supervised semantic maps learned by our \ac{ours}.}
\label{fig:masks}
\end{figure*}
}

\newcommand{\figabmasktwo}{
\begin{figure*}[!tbp]
\centering
\includegraphics[width=1.0\linewidth]{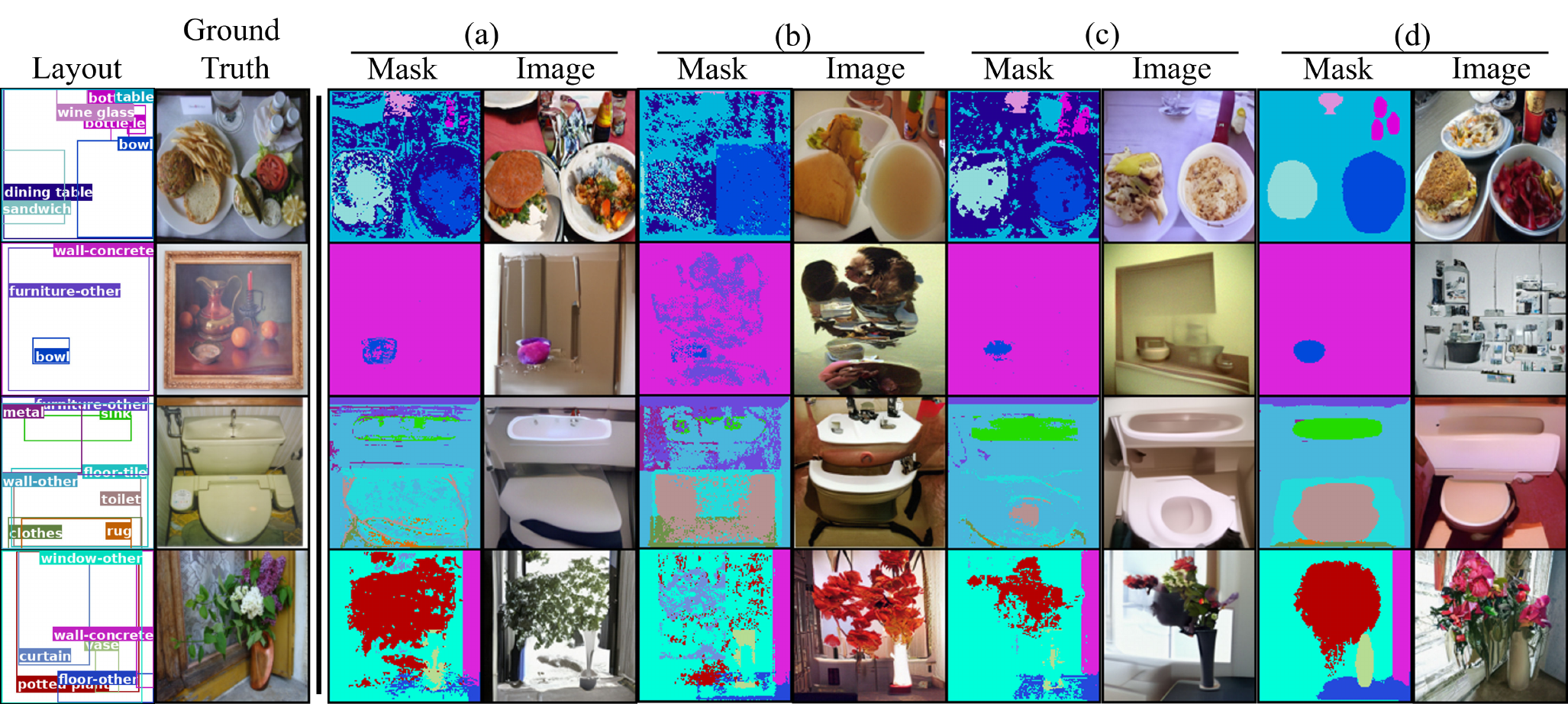}\\
\caption{The generated images and their predicted self-supervised semantic maps from ablated models mentioned in Sec. 4.5. (a) ISLA Norm + Self Attention. (b) \ac{ean} + Self Attention. (c) ISLA Norm + \ac{sma}. (d) \ac{ean} + \ac{sma}.}
\label{fig:abmasktwo}
\end{figure*}
}

\newcommand{\figttoiall}{
\begin{figure*}[!tbp]
\centering
\includegraphics[width=1.0\linewidth]{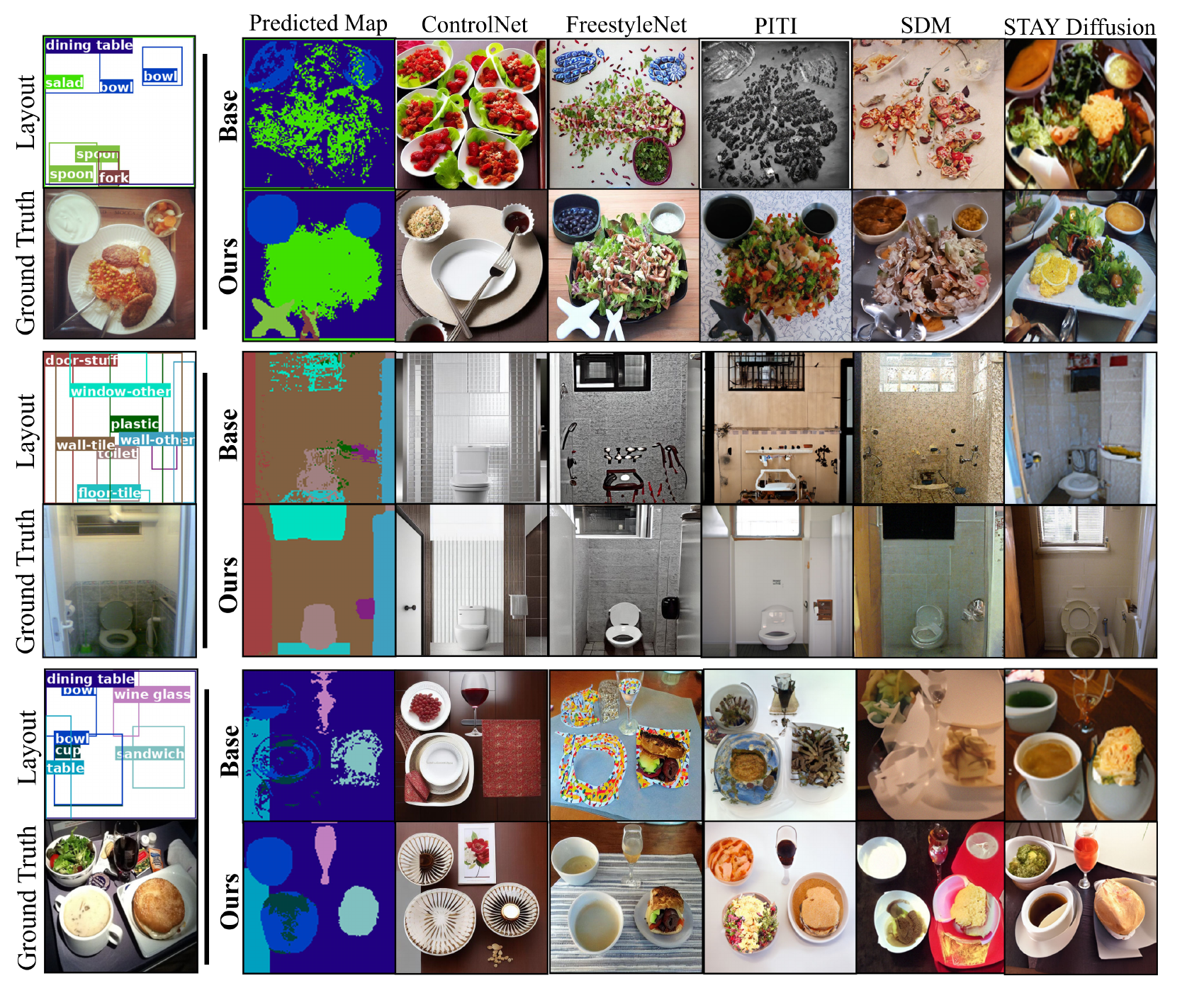}\\
\caption{The visual comparison of images generated from different semantic image synthesis methods and our \ac{ours}. Note that \textbf{Base} indicates the baseline self-supervised map from Tab.~\ref{tab:ablation} setting (a) and \textbf{Ours} indicates the self-supervised map from the full model (i.e., Tab. 2 setting (d)).}
\label{fig:t2iall}
\end{figure*}
}

\newcommand{\tabmetricsrebu}{
\begin{table}[!tbp]
\caption{Quantitative results on COCO-stuff and VG at resolution 256 $\times$ 256. The proposed \ac{ours} outperforms \ac{LayoutDM} in diversity, accuracy and controllability metrics while maintaining close quality performance.}
\vspace{-3.2mm}
\label{tab:metrics_rebu}
\centering
\newcolumntype{x}{>{\centering\arraybackslash\hspace{0pt}}p{13mm}}
\footnotesize{
\begin{tabular}{l@{\hspace{-1.5mm}} l@{\hspace{0.5mm}} |@{\hspace{-2.0mm}} x@{\hspace{-2.0mm}} x@{\hspace{-2.0mm}} x@{\hspace{-4.0mm}} x@{\hspace{-3.5mm}} x@{\hspace{-2.5mm}} x@{\hspace{-0.5mm}} x@{\hspace{-1.5mm}} }
\toprule
 &\multirow{2}{5em}{\textbf{Methods}} &\multicolumn{4}{c}{\textbf{Coco-Stuff}} &\multicolumn{2}{c}{\textbf{VG}}   \\
 \cmidrule(lr){3-6} \cmidrule(lr){7-8}
 & &  FID $\downarrow$   & DS $\uparrow$  & CAS $\uparrow$ & Yolo $\uparrow$  &  FID $\downarrow$   & DS $\uparrow$\\
\midrule \midrule 
&LostGAN v2 \cite{sun2019image} &33.17 &0.55$\pm$0.10 &33.17 &15.0 &34.92 &0.53$\pm$0.01 \\
&PLGAN  \cite{Wang2022InteractiveIS} &30.67  &0.52$\pm$0.10 &29.15 &13.6   &- &- \\
&LAMA \cite{9711206} &33.00  &0.48$\pm$0.12 &9.97 &20.4    &38.51 &0.54$\pm$0.10 \\
&TwFA \cite{Yang2022ModelingIC} &23.78  &0.43$\pm$0.13 &20.09 &23.9   &18.57 &0.50$\pm$0.10 \\
&LayoutDiffuse  \cite{Cheng2023LayoutDiffuseAF} &22.41  &0.58$\pm$0.11 &31.80 &23.7 &22.45 &0.56$\pm$0.10  \\
&GLIGEN \cite{Li2023GLIGENOG} &21.30 &0.57$\pm$0.09 &34.41 &23.0 &23.42 &0.60$\pm$0.09\\ 
&LayoutDM \cite{Zheng_2023_CVPR} &\textbf{15.74}  &0.58$\pm$0.09 &35.69 &27.2 &\textbf{15.26} &0.61$\pm$0.10  \\
&\textbf{Ours} &17.43 &\textbf{0.59$\pm$0.09} &\textbf{37.18} &\textbf{29.5}   &18.02 &\textbf{0.65$\pm$0.08} \\
\bottomrule
\end{tabular}
}
\vspace{-3.5mm}
\end{table}
}

\newcommand{\tabablation}{
\begin{table}[!tbp]
\caption{Ablation study of the proposed \ac{ean} and \ac{sma} modules, where we trained the models on COCO-stuff 128$\times$128.}
\vspace{-3.2mm}
\label{tab:ablation}
\centering
\newcolumntype{x}{>{\centering\arraybackslash\hspace{0pt}}p{14mm}}
\footnotesize{
\begin{tabular}{l@{\hspace{-1.5mm}} l@{\hspace{-0.5mm}} x@{\hspace{-2.5mm}}  x@{\hspace{-0.5mm}} |@{\hspace{-2.5mm}} x@{\hspace{-2.5mm}} x@{\hspace{-0.5mm}} x@{\hspace{-2.5mm}} x@{\hspace{-4.5mm}} x@{\hspace{-2.5mm}}}
\toprule
& &\textbf{Norm} &\textbf{Attention} &  FID $\downarrow$ & IS $\uparrow$  & DS $\uparrow$  & CAS $\uparrow$ & Yolo $\uparrow$  \\
\midrule \midrule 
&(a) &ISLA \cite{sun2019image} &Self &25.44 &13.36$\pm$0.72  &0.56$\pm$0.1 &35.47 &22.10\\
&(b) &EA &Self &24.30 &14.68$\pm$0.56  &\textbf{0.59$\pm$0.1} &33.94 &13.10\\
&(c) &ISLA \cite{sun2019image} &SM &24.60 &15.32$\pm$0.47 &0.57$\pm$0.1 &34.26 &20.90\\
&(d) &EA &SM &\textbf{19.26} &\textbf{15.85$\pm$0.67} &0.57$\pm$0.1 &\textbf{35.94} &\textbf{23.90} \\
\bottomrule
\end{tabular}
}
\vspace{-2.2mm}
\end{table}
}

\newcommand{\tabsis}{
\begin{table}[!tbp]
\caption{Quantitative results of mask clarity on COCO-stuff 256 $\times$ 256. When GT maps are absent, the images generated from the self-supervised maps of our full model (\textbf{Ours}) outperform other baseline methods in quality and controllability metrics, highlighting the importance of clear masks.}
\vspace{-3.2mm}
\label{tab:sis}
\centering
\newcolumntype{x}{>{\centering\arraybackslash\hspace{0pt}}p{14mm}}
\footnotesize{
\begin{tabular}{l@{\hspace{-1.5mm}} l@{\hspace{-3.5mm}} x@{\hspace{-2.5mm}} |@{\hspace{-2.5mm}} x@{\hspace{-2.5mm}} x@{\hspace{0.5mm}} x@{\hspace{-2.5mm}} x@{\hspace{-4.5mm}} x@{\hspace{-3.5mm}} x@{\hspace{-2.5mm}} }
\toprule
 &\textbf{Methods} &\textbf{Mask} &  FID $\downarrow$ & IS $\uparrow$  & DS $\uparrow$  & CAS $\uparrow$ & Yolo $\uparrow$  \\
\midrule \midrule 
&ControlNet~\cite{zhang2023adding} &GT &32.24	&24.80$\pm$1.44	&0.54$\pm$0.08	&22.43	&25.6
 \\
&FreestyleNet~\cite{xue2023freestyle} &GT  &14.80	&30.06$\pm$0.92	&0.50$\pm$0.09	&36.67	&42.9 \\
\midrule
&ControlNet~\cite{zhang2023adding} &Base &64.29	&17.47$\pm$0.49	&0.59$\pm$0.09	&9.36	&4.1\\
&FreestyleNet~\cite{xue2023freestyle} &Base  &37.21	&20.40$\pm$0.73	&0.50$\pm$0.09	&21.54	&12.9
 \\
\midrule
&ControlNet~\cite{zhang2023adding} &Ours &50.25	&20.65$\pm$0.93	&0.60$\pm$0.09	&15.38	&6.5
 \\
&FreestyleNet~\cite{xue2023freestyle} &Ours  &32.95	&22.59$\pm$0.66	&0.56$\pm$0.10	&26.52	&13.8
 \\
&PITI~\cite{wang2022pretraining} &Ours  &37.41	&17.24$\pm$0.91	&0.53$\pm$0.13	&18.63	&12.0
 \\
&SDM~\cite{Wang2022SemanticIS} &Ours  &35.53	&17.70$\pm$0.72	&\textbf{0.69$\pm$0.19}	&22.69	&13.5
 \\
&\textbf{Ours} &Ours &\textbf{17.43} &\textbf{26.08$\pm$0.76} &0.59$\pm$0.09 &\textbf{37.18} &\textbf{29.5} \\
\bottomrule
\end{tabular}
}
\vspace{-4.2mm}
\end{table}
}

\newcommand{\tabmetricsfcoco}{
\begin{table*}[!tbp]
\caption{Quantitative results on COCO-stuff at resolution 256 $\times$ 256. The proposed \ac{ours} outperforms \acf{LayoutDM} in diversity, accuracy and controllability metrics while maintaining close quality performance.}
\vspace{-2.2mm}
\label{tab:metricsfcoco}
\centering
\newcolumntype{x}{>{\centering\arraybackslash\hspace{0pt}}p{18mm}}
\footnotesize{
\begin{tabular}{l l | x x x x x }
\toprule
 &\multirow{2}{5em}{\textbf{Methods}} &\multicolumn{5}{c}{\textbf{Coco-Stuff}}   \\
 \cmidrule(lr){3-7} 
 & &  FID $\downarrow$ & IS $\uparrow$  & DS $\uparrow$  & CAS $\uparrow$ & Yolo $\uparrow$ \\
\midrule \midrule 
&LostGAN v2 \cite{sun2019image} &33.17 &18.08$\pm$0.46 &0.55$\pm$0.10 &33.17 &15.0  \\
&PLGAN \cite{Wang2022InteractiveIS}  &30.67 &18.92$\pm$0.65 &0.52$\pm$0.10 &29.15 &13.6    \\
&LAMA \cite{9711206} &33.00 &19.77$\pm$0.66 &0.48$\pm$0.12 &9.97 &20.4     \\
&TwFA \cite{Yang2022ModelingIC} &23.78 &23.02$\pm$0.94 &0.43$\pm$0.13 &20.09 &23.9   \\
&LayoutDiffuse \cite{Cheng2023LayoutDiffuseAF} &22.41 &27.09$\pm$0.07  &0.58$\pm$0.11 &31.80 &23.7  \\
&GLIGEN \cite{Li2023GLIGENOG} &21.30 &\textbf{27.71$\pm$0.79}  &0.57$\pm$0.09 &34.41 &23.0  \\
&LayoutDM \cite{Zheng_2023_CVPR}  &\textbf{15.74} &26.01$\pm$0.84 &0.58$\pm$0.09 &35.69 &27.2  \\
&\textbf{Ours} &17.43 &26.08$\pm$0.76 &\textbf{0.59$\pm$0.09} &\textbf{37.18} &\textbf{29.5} \\
\bottomrule
\end{tabular}
}
\end{table*}
}

\newcommand{\tabmetricsfvg}{
\begin{table*}[!tbp]
\caption{Quantitative results on \acf{VG} at resolution 256 $\times$ 256. The proposed \ac{ours} outperforms \acf{LayoutDM}  in diversity, accuracy and controllability metrics while maintaining close quality performance.}
\vspace{-2.2mm}
\label{tab:metricsfvg}
\centering
\newcolumntype{x}{>{\centering\arraybackslash\hspace{0pt}}p{18mm}}
\footnotesize{
\begin{tabular}{l l | x x x x }
\toprule
 &\multirow{2}{5em}{\textbf{Methods}} &\multicolumn{4}{c}{\textbf{VG}}   \\
 \cmidrule(lr){3-6} 
 & &  FID $\downarrow$ & IS $\uparrow$  & DS $\uparrow$  & CAS $\uparrow$ \\
\midrule \midrule 
&LostGAN v2 \cite{sun2019image} &34.92 &14.01$\pm$0.81 &0.53$\pm$0.01 &24.40  \\
&PLGAN  \cite{Wang2022InteractiveIS}  &- &- &- &-    \\
&LAMA \cite{9711206}  &38.51 &13.70$\pm$0.76 &0.54$\pm$0.10 &24.16     \\
&TwFA \cite{Yang2022ModelingIC} &18.57 &17.75$\pm$0.68 &0.50$\pm$0.10 &18.30  \\
&LayoutDiffuse  \cite{Cheng2023LayoutDiffuseAF} &22.45 &\textbf{22.89$\pm$1.69}  &0.56$\pm$0.10 &25.05  \\
&GLIGEN \cite{Li2023GLIGENOG} &23.42 &21.84$\pm$1.38  &0.60$\pm$0.09 &25.49  \\
&LayoutDM \cite{Zheng_2023_CVPR}
 &\textbf{15.26} &21.94$\pm$1.28 &0.61$\pm$0.10 &26.84 \\
&\textbf{Ours} &18.02 &18.56$\pm$0.91 &\textbf{0.65$\pm$0.08} &\textbf{27.23} \\
\bottomrule
\end{tabular}
}
\end{table*}
}

\newcommand{\tabsisapp}{
\begin{table*}[!tbp]
\caption{Quantitative results of mask clarity on COCO-stuff. When GT maps are absent, the images generated from the self-supervised maps of our full model (\textbf{Ours}) outperform other baseline methods in quality and controllability metrics, highlighting the importance of clear masks.}
\vspace{-2.2mm}
\label{tab:sisapp}
\centering
\newcolumntype{x}{>{\centering\arraybackslash\hspace{0pt}}p{18mm}}
\footnotesize{
\begin{tabular}{l l@{\hspace{-1.5mm}}  x@{\hspace{-1.5mm}} |@{\hspace{-0.5mm}} x@{\hspace{-1.5mm}} x@{\hspace{0mm}} x@{\hspace{-1.5mm}} x@{\hspace{-3.5mm}} x@{\hspace{-1.5mm}} x@{\hspace{-1.5mm}} }
\toprule
 &\textbf{Methods} &\textbf{Mask} &  FID $\downarrow$ & IS $\uparrow$  & DS $\uparrow$  & CAS $\uparrow$ & Yolo $\uparrow$  \\
\midrule \midrule 
&ControlNet-v1.1  \cite{zhang2023adding} &GT &32.24	&24.80$\pm$1.44	&0.54$\pm$0.08	&22.43	&25.6
 \\
&FreestyleNet \cite{xue2023freestyle} &GT  &14.80	&30.06$\pm$0.92	&0.50$\pm$0.09	&36.67	&42.9 \\
&PITI \cite{wang2022pretraining} &GT &15.22	&28.08$\pm$1.11	&0.45$\pm$0.12	&37.20	&34.8
 \\
&SDM \cite{Wang2022SemanticIS} &GT  &20.79	&23.82$\pm$0.53	&0.65$\pm$0.18	&36.42	&26.9 \\
\midrule
&ControlNet-v1.1 \cite{zhang2023adding} &Base &64.29	&17.47$\pm$0.49	&0.59$\pm$0.09	&9.36	&4.1\\
&FreestyleNet \cite{xue2023freestyle} &Base  &37.21	&20.40$\pm$0.73	&0.50$\pm$0.09	&21.54	&12.9 \\
&PITI \cite{wang2022pretraining} &Base &61.01	&12.90$\pm$0.39	&0.46$\pm$0.12	&14.02	&4.7\\
&SDM \cite{Wang2022SemanticIS} &Base  &40.08	&15.41$\pm$0.26	&0.69$\pm$0.18	&19.29	&10.5 \\
\midrule
&ControlNet-v1.1 \cite{zhang2023adding} &Ours &50.25	&20.65$\pm$0.93	&0.60$\pm$0.09	&15.38	&6.5
 \\
&FreestyleNet \cite{xue2023freestyle} &Ours  &32.95	&22.59$\pm$0.66	&0.56$\pm$0.10	&26.52	&13.8
 \\
&PITI \cite{wang2022pretraining} &Ours  &37.41	&17.24$\pm$0.91	&0.53$\pm$0.13	&18.63	&12.0
 \\
&SDM \cite{Wang2022SemanticIS} &Ours  &35.53	&17.70$\pm$0.72	&\textbf{0.69$\pm$0.19}	&22.69	&13.5
 \\
&\textbf{Ours} &Ours &\textbf{17.43} &\textbf{26.08$\pm$0.76} &0.59$\pm$0.09 &\textbf{37.18} &\textbf{29.5} \\
\bottomrule
\end{tabular}
}
\vspace{-1.2mm}
\end{table*}
}

\newcommand{\tabmetricsgligen}{
\begin{table}[!tbp]
\caption{The reported FID and YoloScore (AP) in Tab.~2 of GLIGEN \cite{Li2023GLIGENOG} and our reproduced results (marked with *).}
\vspace{-3.2mm}
\label{tab:metrics_gligen}
\centering
\newcolumntype{x}{>{\centering\arraybackslash\hspace{0pt}}p{13mm}}
\footnotesize{
\begin{tabular}{l l| x x  }
\toprule
 &{\textbf{Methods}} &FID $\downarrow$    & Yolo $\uparrow$  \\
\midrule \midrule 
&GLIGEN \cite{Li2023GLIGENOG} &21.04 &22.4 \\ 
&GLIGEN \cite{Li2023GLIGENOG}* &21.30 &23.0  \\
\bottomrule
\end{tabular}
}
\vspace{-3.5mm}
\end{table}
}

\usepackage[capitalize]{cleveref}
\crefname{section}{Sec.}{Secs.}
\Crefname{section}{Section}{Sections}
\Crefname{table}{Table}{Tables}
\crefname{table}{Tab.}{Tabs.}

\def\wacvPaperID{645} 
\def\confName{WACV}
\def\confYear{2025}

\begin{document}

\title{STAY Diffusion: \\ Styled Layout Diffusion Model for Diverse Layout-to-Image Generation}

\author{Ruyu Wang\textsuperscript{1,2} \quad Xuefeng Hou\textsuperscript{1} \quad Sabrina Schmedding\textsuperscript{1} \quad Marco F. Huber\textsuperscript{2,3}\\
\textsuperscript{1}Bosch Center for Artificial Intelligence, Renningen, Germany\\
\textsuperscript{2}Institute of Industrial Manufacturing and Management IFF, University of Stuttgart, Stuttgart, Germany \\
\textsuperscript{3}Fraunhofer Institute for Manufacturing Engineering and Automation IPA, Stuttgart, Germany \\
{\tt\small \{ruyu.wang, sabrina.schmedding\}@de.bosch.com \quad houxuefeng1997@gmail.com \quad marco.huber@ieee.org}
}

\maketitle

\begin{abstract}
In \ac{L2I} synthesis, controlled complex scenes are generated from coarse information like  bounding boxes.
Such a task is exciting to many downstream applications because the input layouts offer strong guidance to the generation process while remaining easily reconfigurable by humans. 
In this paper, we proposed \ac{ours}, a diffusion-based model that produces photo-realistic images and provides fine-grained control of stylized objects in scenes. Our approach learns a global condition for each layout, and a self-supervised semantic map for weight modulation using a novel  \ac{ean}. A new \ac{sma} is also introduced to cross-condition the global condition and image feature for capturing the objects' relationships. These measures provide consistent guidance through the model, enabling more accurate and controllable image generation.
Extensive benchmarking demonstrates that our \ac{ours} presents high-quality images while surpassing previous \acl{SOTA} methods in generation diversity, accuracy, and controllability.
\end{abstract}
\vspace{-0.4cm}

\section{Introduction}
\label{sec:intro}
Recently, \acp{DGM} have advanced unprecedentedly on image synthesis tasks \cite{ho2020denoising, song2020denoising, nichol2021improved, dhariwal2021diffusion, song2019generative}. As multiple unconditional generative models \cite{Karras2019stylegan2, ho2020denoising, vahdat2021score} have shown a promising ability to capture data distribution and generate photo-realistic images, conditional generative models aim to achieve the same goal while conditioning the generated images with additional information. 
The given conditional information can range from coarse to detailed, such as class labels, text descriptions,
semantic maps, or even other images, providing various levels of control during the generating process. Such control is favored in many practical applications like graphics editors or data augmentation with specific rare cases in medical or industrial domains. 
However, a trade-off exists between the flexibility for users and the strength of the model control signal: a text description may give little clues for object locations, whereas a semantic map can be hard to acquire and alter. In this respect, using layouts (i.e., a collection of labeled object bounding boxes) as the condition is an appealing solution since a layout offers guidance on object size and locations and is easier to access and reconfigure.

\figteaser
Despite the alluring traits, \acf{L2I} synthesis poses a challenging task. It targets a one-to-many mapping problem involving generating multiple objects, capturing object interactions, and composing perceptually plausible scenes with the given layout. Early works \cite{sun2019image,Ashual2019SpecifyingOA,Zhao2018ImageGF} were primarily 
based on \acp{GAN} due to their outstanding generation ability at the time. However, these models are prone to 
mode collapse and often fail to generate reasonable images with clear objects as illustrated in Fig.~\ref{fig:teaser}(a) (cf. bus in LostGAN v2 \cite{sun2019image}).
Some researchers addressed these problems by changing the network from CNN-based to transformer-based to better capture object relationships of a complex scene \cite{Yang2022ModelingIC}. More recently, the advance of \acp{DM} opened up new possibilities for high-resolution \ac{L2I} synthesis. A few pioneer works have shown new impressive results by tokenizing the given layouts for their new attention mechanisms \cite{rombach2021highresolution, Zheng_2023_CVPR} or by leveraging pretrained \acp{LTGM} \cite{Cheng2023LayoutDiffuseAF, Li2023GLIGENOG}.

In this work, we propose \acf{ours}, a diffusion-based \ac{L2I} model providing fine-grained object control for diverse image synthesis as demonstrated in Fig.~\ref{fig:teaser}(b). Unlike prior works that tokenize \cite{Yang2022ModelingIC, Zheng_2023_CVPR} or apply direct input concatenation to the layout \cite{zhaobo2019layout2im}, we learn a latent representation for each input object to predict pixel-wise object masks based on given layouts. The object representations consist of a category-specific part and an object-specific part that reflects its attribute (i.e., style). Such representations and their corresponding masks are used as the conditions for both the model's normalization and attention layers. We thus refer to them as global conditions. Then, we introduce a new normalization method, termed \acf{ean} to better utilize these global conditions. It embeds the object representations and their masks into the decoder of the diffusion network, where the overlapping areas of object masks are carefully processed for weight modulation. In addition to the \ac{ean}, a novel \acf{sma} mechanism is proposed to cross-condition object representations and image features, thereby enhancing object awareness and capturing object relationships in our model.


Extensive experiments on two benchmarking datasets, COCO-stuff \cite{caesar2018coco} and \ac{VG} \cite{Krishna2016VisualGC}, show that \ac{ours} surpasses previous \ac{SOTA} methods in generation diversity, accuracy, and controllability, demonstrating the effectiveness of our proposed \ac{ean} and \ac{sma}.

Our contributions are summarized as follows: 
\begin{itemize}
  \item We propose a novel framework called \acl{ours}, a diffusion-based generative model for synthesizing high-fidelity and diverse images with precise control over multiple objects in complex scenes.
  \item Unlike other methods, we process the layouts into the global conditions (i.e., object representations and their predicted masks) and introduce two new modules---\acl{ean} and \acl{sma}---utilizing the global conditions for fine-grained object control and scene understanding.
  \item  The proposed \ac{ours} produces high-quality images together with their self-supervised maps, while outperforming the previous \ac{SOTA} methods regarding diversity, accuracy, and controllability in generation. 
\end{itemize}

\section{Related Works}

\textbf{Conditional Image Synthesis.}
Since the breakthrough in high-resolution photo-realistic image synthesis made by \acp{GAN} \cite{goodfellow2014generative}, \acp{DGM} have attracted growing attention. While unconditional synthesis models \cite{zhang2019self, Karras2019stylegan2, ho2020denoising, song2020denoising, nichol2021improved} only take random noise as input and mainly focus on improving generation quality and capturing the underlying data distribution, conditional synthesis models \cite{Mirza2014ConditionalGA, brock2018large, dhariwal2021diffusion, ho2022classifier, Radford2021LearningTV, Nichol2021GLIDETP, rombach2021highresolution} consider additional input signals (e.g., category, text, image) to achieve higher controllability along the generation process. Various ways to embed conditional signals into \acp{DGM} have been studied. For example, some works \cite{Odena2016ConditionalIS,Reed2016GenerativeAT} proposed to concatenate the conditions naively with input or intermediate network features, while others \cite{sun2019image, He2021ContextAwareLT, Wang2022InteractiveIS,9711206} suggested injecting them in a projection-based way \cite{Miyato2018cGANsWP}. A popular trend is to modulate the normalization layers in models with conditional gains and bias \cite{Huang2017ArbitraryST, park2019SPADE, schonfeld2021you}, which are commonly used in semantic image synthesis. Our method is closely related to this approach, but instead of a semantic map, a layout as input is sufficient.
Most early works conducted their investigations in \acp{GAN} due to its superior performance at the time. Recently, \acp{DM} have emerged as the new \ac{SOTA} family in image synthesis, where the \ac{ADM} \cite{dhariwal2021diffusion} first showed 
the promising prospect of \acp{DM} by outperforming \ac{GAN}-based methods. Since then, many researchers have explored different ways such as guiding the generation process by conditional gradients \cite{ho2022classifier} to condition \acp{DM} on additional information \cite{rombach2021highresolution, Avrahami_2022_CVPR, LI202247,Lugmayr2022RePaintIU, Wang2022SemanticIS, wang2022pretraining}. Beside training a conditional \acp{DM} from scratch, leveraging pretrained \acp{LTGM}  \cite{rombach2021highresolution,saharia2022photorealistic, mishkin2022risks} for conditional image generation \cite{mou2024t2i, zhang2023adding, xue2023freestyle} is also a popular trend due to the resource efficiency and the rich expressiveness offered by large models.


\textbf{Layout-to-Image Synthesis.}
\ac{L2I} synthesis is one of the challenging tasks in conditional image synthesis, where only coarse information (e.g., category, size, and location) of objects is given to compose a complex scene. Layouts, consisting of multiple labeled bounding boxes, were often used to aid other tasks like text-to-image synthesis prior to Layout2Im \cite{Zhao2018ImageGF}. 
Attracted by the offered flexibility and controllability, researchers have explored many ways to adapt layouts into \acp{DGM}: LostGAN v2 \cite{sun2019image} encode layouts into style features and object masks for ISLA Norm modulation. 
The following works \cite{Wang2022InteractiveIS,9711206, He2021ContextAwareLT} employed a similar mechanism to embed the layout information. Moreover, they introduced additional designs to tackle the object interaction and mask clarity for better generation quality.
Apart from \ac{GAN}-based models, Taming \cite{Jahn2021HighResolutionCS} and TwFA \cite{Yang2022ModelingIC} used a pretrained VQ-GAN \cite{Esser2020TamingTF} as layout encoder and trained auto-regressive transformers on top to perform image generation.
As a multimodal model, LDM \cite{rombach2021highresolution} was the first \ac{DM} to support \ac{L2I} synthesis
, demonstrating great potential to utilize \ac{DM} as the backbone for such a task. 
More recently, three more dedicated \acp{DM}, \ac{LayoutDM} \cite{Zheng_2023_CVPR},  LayoutDiffuse \cite{Cheng2023LayoutDiffuseAF} and GLIGEN \cite{Li2023GLIGENOG}, were proposed. The first one introduced a novel module to better fuse the tokenized inputs (i.e., image and layout) for conditioning the model, while the other two leveraged a pretrained \ac{LTGM} \cite{rombach2021highresolution} for \ac{L2I} generation to reduce training workload and exploit the rich semantic learned from large datasets. 
Our \ac{ours} does not deploy pretrained \acp{LTGM} as in LayoutDiffuse or GLIGEN, allowing for higher adaptability when applying to domains beyond natural (e.g., medical or industrial) images.
Instead of tokenizing layouts with additional models like in \ac{LayoutDM}, we learn object representations and masks from layouts. The information is then treated as global conditions for both normalization and attention layers in our work.


\section{Methodology}
\label{sec:method}
In this section, we present a novel framework named \ac{ours} to transform coarse layout information into realistic images. Specifically, we refer to coarse layout information as a collection of bounding boxes and their categorical labels, which provides a rough sketch of a complex scene. 
Through the iterative diffusion steps, the proposed \ac{ours} generates high-quality images from layouts. In the following, we begin with the model overview. Then, two crucial designs, \acl{ean} and \acl{sma}, are introduced. Finally, we present the training and sampling methods for our framework. 

\figmodel

\subsection{Layout-Conditional Diffusion Model}
The overview of our model is illustrated in Fig.~\ref{fig:model}. A layout $l$ is defined as a set of $N$ objects, where each object is represented as $o_i \! = \! \{b_i, c_i\}$, $i = 1,\ldots, N$. 
Precisely,  $b_i \! = \! (x_0^i, y_0^i, h, w) \! \in \! [0,1]^4 $ denotes a bounding box in a normalized coordinate system and $c_i \in \{1, \ldots, C\}$ is the object category.
To account for the case where the number of objects in each layout is different, we introduce a padding class $c=0$ and set the coordinates as $b_0=(0,0,0,0)$ to pad the number of objects in each layout to a predefined number.
Consider a data point $x_0$ sampled from a real data distribution $q(\cdot)$ and a layout $l$ as a condition: the goal of a conditional \ac{DM} is to maximize the likelihood $p_\theta(x_0|l)$ following the conditional data distribution $q(x_0|l)$.  
Two processes---the forward process and the reverse process---are defined to achieve the desired data generation. 
Starting from $x_0$, the forward process $q(x_{1:T}|x_0)$ gradually adds a small amount of Gaussian noise at each timestep $t$ according to
\begin{align}
    q(x_t | x_{t-1}) = \mathcal{N}(x_t; \sqrt{\alpha_t} \cdot x_{t-1}, (1-\alpha_t)\cdot\mathbf{I})~,
    \label{eq:forward}
\end{align}
where $1-\alpha_t$ represents the noise magnitude. Assuming the added noise is from a diagonal Gaussian distribution and a sufficient amount is applied through the forward process, the acquired $x_T$ can be approximated by  $\mathcal{N}(0, \mathbf{I})$.
On the other hand, the reverse process $p_\theta(x_{0:T}|l)$ uses a parameterized network to estimate the real posterior $q(x_{t-1}|x_t)$
by 
\begin{align}
    p_{\theta}(x_{t-1}|x_t, l) = \mathcal{N}(x_{t-1}; \mu_{\theta}(x_t,l,t), \Sigma_{\theta}(x_t,l,t))~,
    \label{eq:backward}
\end{align}
progressively predicting a less noisy image at each step to recover $x_0$.
Following Ho et al. \cite{ho2020denoising}, the optimization objective of the conditional \ac{DM} can be simplified as 
\begin{equation}
    \mathcal{L(\theta)} \! = \! \mathrm E_{t , x_0, l, \epsilon}\big[||\epsilon - \epsilon_{\theta}(x_t, l, t)||^2\big]~,
    \label{eq:lsimple}
\end{equation}
where $\epsilon$ is the added noise in the forward process at each timestep $t$ and $\epsilon_{\theta}$ is its approximation predicted by the model in the backward process. The goal is to only predict the added noise for removal instead of the entire image.

We based our model on \ac{ADM}, where a U-Net \cite{ronneberger2015u}-based network is deployed for the denoising process.
In addition to \ac{ADM}, we introduce learnable object representations and a mask prediction subnetwork to process the layout information into unified global conditions for our model. Specifically, an object representation $o^\mathrm{sty} \in \mathbb{R}^{1 \times (d_{e}+d_{z})}$ is a concatenation of a learnable categorical embedding $c^\mathrm{emb} \in \mathbb{R}^{1 \times d_{e}}$ and a randomly sampled  $z^\mathrm{sty} \in \mathbb{R}^{1 \times d_{z}} \sim \mathcal{N}(0,1)$ for controlling the object-specific details (i.e., style). 
The mask prediction subnetwork consists of a series of convolution layers, followed by a sigmoid transformation. It takes $o^\mathrm{sty}$ and predicts its initial object probabilistic mask $M^0$. 
Then, an input layout $l$ is a collection of objects $\{o^\mathrm{sty}_i\}^{N}_{i=1}$ and their respective masks $\{M^0_i\}^{N}_{i=1}$, which we refer to as the global conditions for the \ac{DM}.

As for the U-Net, we inherit the encoder part of \ac{ADM} for noisy image encoding while designing novel modules for the decoder part to embed the global conditions. 
Particularly, we propose to use Layout Diffusion ResBlocks for the decoder part of our model, where we replace the Group Normalization \cite{wu2018group} layers with novel \acf{ean} layers, and the updated object masks $M^{j+1}$ are predicted at the end of the block (cf. Fig.~\ref{fig:blockss}(a) in Appendix~\ref{app:imp_blocks}). 
Note that before passing to the next block, we follow the design in \cite{sun2019image} to further refine $M^{j+1}_i$ with $M^{j+1}_i = (1-\eta)\cdot M^0_i + \eta \cdot M^{j+1}_i$, where $\eta$ is a learnable weight and $i$, $j$ are the object and ResBlock indices, respectively. The proposed \ac{ean} is designed to fuse multiple objects smoothly and enable spatially-adaptive weight modulation.
Additionally, a novel \ac{sma} is employed, allowing the model to only attend to relevant regions based on the given object masks.

\subsection{\acl{ean}}
\label{sec:ean}
Inspired by works in semantic image synthesis \cite{park2019SPADE, schonfeld2021you, Wang2022SemanticIS}, the proposed \ac{ean} modulates the weight and bias of a normalization layer with a semantic map. 
Instead of using precisely labeled maps as input, where each pixel belongs to only one object, our \ac{ean} is designed to work with self-supervised object masks predicted from the layouts. 
Specifically, we noticed that methods that predict object masks independently, like \cite{sun2019image} or ours, create false overlapping between objects (cf. $B^{j}$ in Fig.~\ref{fig:eanorm}).
This results in multiple modulations being applied to one pixel.
We hypothesize that ensuring the purity of these masks (i.e., the certainty of a pixel's class) can significantly improve the quality of the assembled self-supervised map along with its generated image (cf. Sec.~\ref{sec:mask_clear} for empirical results on this hypothesis).
Thus, we design a new way to carefully utilize the self-supervised masks while limiting their overlap (see paragraph Edge-Aware Weighted Map below).
Moreover, we propose to extend each predicted pixel with its object representation $o^\mathrm{sty}_i$ to improve the map's expressiveness and the object controllability for guiding the generation.
We illustrate the whole workflow in Fig.~\ref{fig:eanorm} and formulate it as
\begin{equation}
    \mathbf{feat}^{j+1} = \bm{\gamma}^j \cdot \mathrm{BatchNorm}(\mathbf{feat}^j) + \bm{\beta}^j~,
\end{equation}
where $\mathbf{feat}^{j}$ and $\mathbf{feat}^{j+1}$ are the input and output features of \ac{ean}, respectively. The spatially aware weight $\bm{\gamma}^j$ and bias $\bm{\beta}^j$ are learned from the layout as follows.
\figeanorm

\textbf{Edge-Aware Weighted Map.}
To improve the clarity of object masks, we construct a pixel-level weighted semantic map from the object masks based on a given layout:
we place the probabilistic object masks $M^j$ to their bounding box locations and turn them into binary masks $B^j$ by thresholding at $0$.
A \textbf{Non-Overlapping} semantic map $m^\mathrm{non}$ is computed such that for each pixel location only the object with the smallest mask in $B^j$ remains activated (i.e., remains 1); otherwise, the pixel value is reset to 0. This is because we assume smaller objects have higher priority in claiming a pixel than larger ones.
However, using the non-overlapping map alone may lead to unnatural boundaries in the generated images. Therefore, we also compute an \textbf{Edge-Aware} semantic map $m^\mathrm{ea}$, which extends each object mask by one pixel at the borders using a mask dilation \cite{enwiki:1182952460} operation. We associate the $m^\mathrm{ea}$  with learnable weights $w$ for each object to fuse multiple objects in a given layout adaptively in the next step. 
The pixel-wise weighted map 
 $W^j_i$ for object $i$ is then constructed as 
\begin{equation}
    \mathbf{W}^{j}_{i} =(m^\mathrm{non}_i \odot M^j_i)  + \alpha \cdot (m^\mathrm{ea}_i \cdot w_i)~,
\end{equation}
where $\odot$ stands for the element-wise multiplication, $i$ denotes the object index, $w_i$ is the learnable edge weight to tune the intensity of each $m^\mathrm{ea}$, and $\alpha$ is a hyperparameter to balance between the two semantic maps. 
Finally, the object- and style-aware map $\widehat{W}^j$ is computed as the sum of the object representations $o^\mathrm{sty}$ at each pixel location, weighted by the Edge-Aware Weighted Map $W^j$.
The final spatially aware $\bm{\gamma}^j$ and $\bm{\beta}^j$ are then acquired via applying linear projection to $\widehat{W}^j$.

\subsection{Styled-Mask Attention}
\label{sec:sma}
To further strengthen the network's attention on the given conditional information, we propose the \ac{sma} module, in which we replace the original self-attention by a novel mask attention layer (cf. Fig.~\ref{fig:blockss}(b) in Appendix~\ref{app:imp_blocks}). A standard QKV attention \cite{vaswani2017attention} is defined as $\mathrm{Attention}(Q, K, V) = \mathrm{softmax}(QK^T/\sqrt{d})V$, where $Q$, $K$, and $V$ represent query, key, and value embeddings, respectively. 
Our \ac{sma} uses a cross-attention mechanism to stress the module to only pay attention to localized object regions according to a given layout. 
Specifically, we construct our query $Q^j$ at layer $j$ from the global conditions as follows: First, we compute a pixel-wise semantic map $\widehat{m}^\mathrm{attn} \in \mathbb{R}^{h \times w \times 1}$ with
\begin{equation}
    \widehat{m}^\mathrm{attn}(x,y) = \mathrm{argmax}_{i}(\psi(M^{j}_i(x,y)))~,
\end{equation}
where $x$, $y$ are the map's pixel coordinates, $i$ is the object index, $M^j$ are the object masks that are placed to their corresponding location and $\psi(\cdot)$ denotes the operation to remove the overlapping pixels between objects. 
Then, we compute a more expressive semantic map $m^\mathrm{attn} \in \mathbb{R}^{h \times w \times (d_e+d_z)}$ based on $\widehat{m}^\mathrm{attn}$ by
\begin{equation}
    m^\mathrm{attn}(x,y) = o^\mathrm{sty}_{\widehat{m}^\mathrm{attn}(x,y)} \cdot M^{j}_{\widehat{m}^\mathrm{attn}(x,y)}(x,y)~,
\end{equation}
where  $o^\mathrm{sty}_{\widehat{m}^\mathrm{attn}(x,y)}$ is the object representation of the predicted class in $\widehat{m}^\mathrm{attn}$ and $M^{j}_{\widehat{m}^\mathrm{attn}(x,y)}(x,y)$ is its predicted probability.
Finally, the $Q$, $K$ and $V$ values for our mask attention layer are acquired by  
$Q^j = \varphi_Q(m^\mathrm{attn})~, K^j = \varphi_K(\mathbf{feat}^j)~, V^j = \varphi_V(\mathbf{feat}^j)$, where $\varphi_{Q},\varphi_{K},\varphi_{V}$ are linear projection layers. 

\subsection{Training and Sampling Schemes}
Our \ac{ours} is trained with two objective functions: $\mathcal{L}_\mathrm{simple}$ and $\mathcal{L}_\mathrm{vlb}$ as in \ac{ADM}. Given an image $x$ and a random timestep $t \in \{0,1,...,T\}$, a noisy image $\widetilde{x}$ can be acquired through Eq.~\eqref{eq:forward}. The goal of our conditional \ac{DM} is to reconstruct $x$ by removing the added noise $\epsilon$ at the timestep $t$ under the guidance of a given layout $l$. This objective is referred to  as $\mathcal{L}_\mathrm{simple}$, which is formulated in Eq.~\eqref{eq:lsimple}. Additionally, we follow the proposal of Nicalol and Dhariwal \cite{nichol2021improved} to not only model the noise but also parameterize the variance $\Sigma_\theta(\widetilde{x}_t, l, t)$. 
The associated loss term $\mathcal{L}_\mathrm{vlb}$ is then defined as 
\begin{equation}
    \mathcal{L}_\mathrm{vlb} = \mathrm{KL}( q(\mathbf{x}_{t-1}|\mathbf{x}_t, \mathbf{x}_0 || p_\theta(\mathbf{x}_{t-1}|\mathbf{x}_t, \mathbf{l})))
\end{equation}
and the overall objective is the weighted sum 
\begin{equation}
    \mathcal{L} = \mathcal{L}_\mathrm{simple} + \lambda \cdot \mathcal{L}_\mathrm{vlb}~,
\end{equation}
where $\lambda$ is the non-negative trade-off hyperparameter to balance loss functions.

However, in contrast to \ac{ADM}, we do not train an additional classifier to support the conditional sampling at inference time. Instead, we adapt the classifier-free guidance proposed in \cite{ho2022classifier} to guide the sampling processes. The classifier-free guidance is accomplished by interpolating between the conditioned and unconditioned outputs of the model, where the unconditioned output can be acquired by replacing the given layout $l$ with an empty layout $l_{\emptyset} = \{(0, b_0)_i\}^N_{i=1}$. 
At inference,  the sampling procedure
\begin{equation}
    \hat{\epsilon}_\theta(x_t,l,t) = \epsilon_\theta(x_t,l_{\emptyset},t) + s \cdot \left(\epsilon_\theta(x_t,l,t) - \epsilon_\theta(x_t,l_{\emptyset},t)\right)
\end{equation}
is used, where $\epsilon_\theta$ is the predicted noise from Eq.~\eqref{eq:lsimple} and the non-negative scale $s$ is a hyperparameter used to tune the strength of the guidance.

\section{Experiments}
In this section, we evaluate our \ac{ours} on different benchmarks in terms of various metrics. First, we introduce the experimental settings including datasets, evaluation protocols and metrics. Second, we present the qualitative and quantitative comparisons between our method and previous \ac{SOTA} methods. 
Third, we provide ablation studies to validate the effectiveness of the proposed \ac{ean} and \ac{sma} modules.
Finally, we investigate the importance of the clarity of our self-supervised maps.

\subsection{Datasets}
Two benchmark datasets in \ac{L2I} synthesis, \textbf{COCO-Stuff} \cite{caesar2018coco} and \textbf{\acf{VG}} \cite{Krishna2016VisualGC}, are used for our evaluations.
The COCO-Stuff dataset contains pixel-level annotations for 80 things and 91 stuff classes. 
Following prior works \cite{sun2019image, Zheng_2023_CVPR}, we narrowed the image object count to 3--8 and excluded those with less than 2\% coverage, yielding 112,680 training and 3,097 testing images across 171 categories.
The \ac{VG} dataset presents 108,077 images with dense annotation of objects, attributes, and relationships. We removed small objects and selected images of 3--30 bounding boxes as in \cite{sun2019image, Zheng_2023_CVPR}. The refined dataset has 62,565 training and 5,096 testing images from 178 categories.  

\subsection{Evaluation Protocol  \& Metrics}
\label{eval-metric}
We compared our results to previous \ac{L2I} synthesis methods---LostGAN v2 \cite{sun2019image}, PLGAN \cite{Wang2022InteractiveIS}, LAMA \cite{9711206}, TwFA \cite{Yang2022ModelingIC}, LayoutDiffuse \cite{Cheng2023LayoutDiffuseAF}, GLIGEN \cite{Li2023GLIGENOG}, and \ac{LayoutDM}  \cite{Zheng_2023_CVPR}---, and evaluated our self-supervised maps via semantic image synthesis benchmarks---ControlNet-v1.1 \cite{zhang2023adding}, FreestyleNet \cite{xue2023freestyle}, PITI \cite{wang2022pretraining}, and SDM \cite{Wang2022SemanticIS}. 
All the following evaluations were done with models trained at resolution 256 $\times$ 256 unless otherwise specified. 
See Appendix~\ref{app:imp_details} for how we acquired images from other methods and more details of \ac{ours}, including model architecture, training, and sampling hyperparameters.

Five popular metrics were used to evaluate the generated images from multiple aspects: we chose Fréchet inception distance (FID) \cite{Heusel2017GANsTB} and Inception Score (IS) \cite{salimans2016improved} for measuring the overall quality and visual appearance of the images. 
As for Diversity Score (DS), which estimates the diversity of the generated images, we follow other works \cite{sun2019image, Zheng_2023_CVPR, Cheng2023LayoutDiffuseAF} and adapt LPIPS \cite{zhang2018perceptual} to compute the distance between two images from the same layout in the feature space. 
Also, we demonstrate the controllability of our method by calculating the Classification Score (CAS) \cite{ravuri2019classification}, and the YoloScore (Yolo) \cite{9711206}. The former uses the cropped generated images to train a ResNet-101 \cite{he2016deep} classifier and test it on real images; the latter deploys a pretrained YoloV4 \cite{bochkovskiy2020yolov4} model to detect the 80 thing categories from the generated images. In other words, the model must produce high-quality recognizable objects within the designated bounding boxes to reach a high CAS and YoloScore. 

\figqual

\subsection{Qualitative Results}
\label{sec:qual}

We present the qualitative comparisons of our \ac{ours} to previous methods on COCO-Stuff in Fig.~\ref{fig:qual}. Compared to other methods, \ac{ours} produces more accurate high-quality images following the given layouts, while others like LayoutDiffuse produce artifacts despite using \ac{LTGM}. Especially in scenes with complex object relationships, our method generates images with recognizable objects in the bounding boxes and has the least distortion. For example, in the first row of Fig.~\ref{fig:qual}, only \ac{ours} produced recognizable persons within the four bounding boxes labeled as ``people''. Also, in the last row, our image faithfully presents the effect of fog  while methods like \ac{LayoutDM} failed. 
Moreover, the design of our \ac{ean} module allows \ac{ours} to achieve fine-grained controllability over object appearance and predict self-supervised maps for the generated images. As shown in Fig.~\ref{fig:stya}, \ac{ours} can change the object's attribute (i.e., the shape of the rock) by resampling its associated $z^\mathrm{sty}$ while conditioning on the same layout and input noise. This feature is especially appealing in domains like medical or industrial image augmentation, where a slight difference in object appearance (e.g., disease or defects) matters. On the other hand, \ac{LayoutDM} does not offer such kind of control. 

In Fig.~\ref{fig:maskb}, we present the self-supervised semantic maps learned in the \ac{ean} based on the provided layouts and illustrate the output diversity when the model is conditioned on the same layout but with different input noises (cf. Appendix~\ref{app:more_images} for more results). Additionally, we showcase the interactivity of our approach in Appendix~\ref{app:interact}.

\figsty
\tabmetricsrebu

\subsection{Quantitative Results}
\label{sec:quant}
We followed the protocol in \cite{Zheng_2023_CVPR} to form the evaluating sets by randomly sampling five images for each layout in COCO-stuff and one for \ac{VG} for a fair comparison.  The quantitative results of our method in comparison to previous works are presented in Tab.~\ref{tab:metrics_rebu}. Due to the space limit, please see Appendix~\ref{app:more_tables} for the full tables. 
It can be observed that our \ac{ours} surpasses non-\ac{DM} methods in all metrics---notably, also the \ac{LTGM}-based  LayoutDiffuse and GLIGEN.
As for the \ac{SOTA} \ac{LayoutDM}, our method outperforms in DS, CAS, and YoloScore while maintaining close performance in FID.
We hypothesize that the differences in FID are due to the tighter constraints introduced by our proposed \ac{ean} and \ac{sma}, which heavily rely on the self-supervised maps. On the one hand, this allows superior accuracy and controllability in object generation as shown in Sec.~\ref{sec:qual}. On the other hand, imperfections in the self-supervised maps can pose a hazard to the model and hinder the image fidelity. We further study the impact of the map's quality in Sec.~\ref{sec:mask_clear}.



\figablation
\tabablation
\subsection{Ablation Study}

We conducted ablation experiments on different variants of our proposed method to understand the contributions of each module. In Tab.~\ref{tab:ablation}, we present four settings. We considered the original \ac{ADM} with a self-attention mechanism as the foundation and adapted ISLA Norm from LostGAN v2, which allows object overlapping, to take the layout as the condition. This case is referred to as (a) and is treated as the baseline for our method. Based on setting (a), we changed one component at a time in settings (b) and (c), where our \ac{ean} replaced the ISLA Norm in (b), and the self-attention was replaced by our \ac{sma} in (c), respectively. Finally, setting (d) stands for our full model. We evaluated all four variants with the evaluation metrics in Sec.~\ref{eval-metric}. The improvement in FID, IS, and DS shows that the use of \ac{ean} and \ac{sma} allows for higher generation quality and diversity. However, it can be observed that the performance of CAS and YoloScore fluctuated in settings (b) and (c). We hypothesized that it is due to the lack of global conditional information: the object awareness strengthened by the \ac{ean} was lost in the self-attention mechanism, and the ISLA Norm did not provide a strong localization signal for the \ac{sma} to follow. Our final model, which combines \ac{ean} with \ac{sma}, does not suffer from this issue. Moreover, it provides a better balance between diversity (i.e., DS) and controllability (i.e., CAS and YoloScore) compared to setting (a), suggesting the importance of the global condition. 

Apart from the quantitative evaluation, we present the generated images and self-supervised semantic maps of all settings in Fig.~\ref{fig:ablation}. It can be seen that when the \ac{ean} is applied (i.e., in settings (b) and (d)), the predicted masks for objects are more integrated, demonstrating the effectiveness of our \ac{ean}. ISLA Norm (i.e., in settings (a) and (c)), on the other hand, produces masks with fragmented predictions. More results can be found in Appendix~\ref{app:more_images}.

\subsection{Mask Clarity}
\label{sec:mask_clear}
To investigate the importance of mask clarity as we hypothesized in Sec.~\ref{sec:ean}, we provided three types of masks: the ground truth map (\textbf{GT}), the self-supervised map with overlapping objects from Tab.~\ref{tab:ablation}(a) (\textbf{Base}) and the self-supervised map from our full model (\textbf{Ours}) to the \ac{SOTA} semantic synthesis methods for image generation. The evaluation results are shown in Tab.~\ref{tab:sis} (cf. Appendix~\ref{app:more_clarity} for the full table and visual comparison).
It can be observed that methods using GT maps show the best performance in all cases. However, if the perfectly labeled semantic maps are unavailable, our method and its self-supervised maps outperform all baselines in quality and controllability metrics. 
This showcases the superiority of our method in the case where only a bounding box layout is available. 
Moreover, we can observe a clear trend in method performance: \textbf{GT} map is better than \textbf{Ours}, which again is better than \textbf{Base}. This indicates that further improving self-supervised maps' quality (e.g., through increasing mask clarity) is a promising way to enhance generation quality. We leave this for future work.

\tabsis

\section{Conclusion}
In this paper, we introduce \acf{ours}, a diffusion-based model with novel designed normalization and attention modules for layout-to-image synthesis. With the newly proposed \acl{ean} module, our diffusion model learns a latent representation for each bounding box object in a layout and predicts a self-supervised map for more precise condition guidance. In addition, we proposed the \acl{sma} to capturing the object's relationship based on the learned object representations and maps from the normalization module, forming consistent global conditions through the model. These two critical designs allow our \ac{ours} to perform more fine-grained object control compared to previous methods. 
Extensive results on the challenging COCO-stuff and Visual Genome dataset show that STAY Diffusion can not only produce high-quality images but also outperform previous \acl{SOTA} regarding generation diversity, accuracy, and controllability.


{\small

\bibliographystyle{ieee_fullname}
\bibliography{PaperForReview}
}
\clearpage
\appendix
\setcounter{table}{3}
\setcounter{figure}{7}


\section{Implementation and Evaluation Details}
\label{app:imp_details}
\subsection{Design of \ac{ours}'s Network Blocks}
\label{app:imp_blocks}
We illustrate the design of the Layout Diffusion Resblock and the Diffusion StyledMaskAttnBlock in Fig.~\ref{fig:blockss}. 
In the Layout Diffusion Resblock (see Fig.~\ref{fig:blockss}(a)), we replace the Group Normalization \cite{wu2018group} layers with novel \acf{ean} layers, and the updated object masks $M^{j+1}$ are predicted at the end of the block. Before passing to the next block, we follow the design in \cite{sun2019image} to further refine $M^{j+1}_i$ with $M^{j+1}_i = (1-\eta)\cdot M^0_i + \eta \cdot M^{j+1}_i$, where $\eta$ is a learnable weight and $i$, $j$ are the object and ResBlock indices, respectively. Note that the $M^{j}$ used for the current Resblock is the weighted sum of the initial object masks $M^{0}$ and the predicted masks $M^{j}$ from the previous Resblock. As for the Diffusion StyledMaskAttnBlock, we replace the original self-attention by the proposed \acf{sma} layer as shown in Fig.~\ref{fig:blockss}(b).

\figblockss

\subsection{Evaluation Images of Compared Methods}
\label{app:imp_imgothers}
For \acf{L2I} synthesis methods, we acquired images of LostGAN v2 \cite{sun2019image}, PLGAN \cite{Wang2022InteractiveIS}, LAMA \cite{9711206}, TwFA \cite{Yang2022ModelingIC}, LayoutDiffuse \cite{Cheng2023LayoutDiffuseAF}, and \acf{LayoutDM}  \cite{Zheng_2023_CVPR} by sampling from the checkpoints released by the authors. Since there is no publicly available COCO-stuff checkpoint for GLIGEN\cite{Li2023GLIGENOG}, we followed the setting described by the authors to train a model on COCO-stuff based on LDM\cite{rombach2021highresolution}. We then sampled from this model, following the instructions of the authors such as using scheduled sampling. Our evaluation results on the sampled images (cf. Tab.~\ref{tab:metrics_rebu} and Tab.~\ref{tab:metricsfcoco}) closely matched the reported numbers in GLIGEN (cf. Tab. 2\footnote{The authors of GLIGEN refer to COCO-stuff as COCO2017D.} in their main paper). We also provide a quick comparison in Tab.~\ref{tab:metrics_gligen}.
As for \acf{VG}, we sampled from the checkpoint provided by the authors of GLIGEN. 

\tabmetricsgligen

For semantic image synthesis benchmark methods, we selected ControlNet-v1.1 \cite{zhang2023adding}, FreestyleNet \cite{xue2023freestyle}, PITI \cite{wang2022pretraining} and SDM \cite{Wang2022SemanticIS} for evaluation. We used the checkpoints released by the authors and gave the models different sets of semantic maps (i.e., the ground truth map (\textbf{GT}), the self-supervised map with overlapping objects from Tab.~\ref{tab:ablation}(a) (\textbf{Base}) and the self-supervised map from our full model (\textbf{Ours})) to generate images for evaluation.

\subsection{Training and Sampling Details for \ac{ours}}
\label{app:imp_hyperparam}
We reported the used hyperparameters of \ac{ours} for training and sampling in Tab.~\ref{tab:trainhyper}. For models trained at resolution 256 $\times$ 256, we used four Tesla A100 for training. For models trained at resolution 128 $\times$ 128, we used four Tesla V100 for training. Finally, a Tesla V100 was used to sample images from both resolutions.

\figcombadd
\section{Interactivity of \ac{ours}}
\label{app:interact}
We demonstrate the interactivity of \ac{ours} in Fig.~\ref{fig:combadd}. Although imperfect, the self-supervised maps generated by \ac{ours} can help reduce human effort for image labeling or provide more comprehensive information for downstream tasks such as image blending. As shown in Fig.~\ref{fig:combadd}(a), the mask extracted from \ac{ours} is more accurately aligned with the object shape than the one drawn from a raw bounding box. Furthermore, in Fig.~\ref{fig:combadd}(b), we gradually added an object to the layout to demonstrate that \ac{ours} can be easily adjusted to reconfigurations.

\section{Additional Results}
\label{app:more}
\subsection{Full Quantitative Results}
\label{app:more_tables}
We reported the full quantitative results on COCO-stuff in Tab.~\ref{tab:metricsfcoco} and \acf{VG} in Tab.~\ref{tab:metricsfvg}. Note that the YOLOScore is only applicable for COCO-stuff as defined in \cite{9711206}. As shown in both tables, our \ac{ours} presents superior performance in image diversity, generation accuracy, and controllability. As for image quality, our method shows comparable results to the previous \acf{SOTA} in FID and IS.

\subsection{More Visualization Results}
\label{app:more_images}
We show more visual comparison to previous methods on COCO-stuff in Fig.~\ref{fig:qualtwo} and Fig.~\ref{fig:qualthree}, and \ac{VG} in Fig.~\ref{fig:vgqualone} and Fig.~\ref{fig:vgqualtwo}. Additionally, we provided more demonstration of style variations on object appearance in Fig.~\ref{fig:stytwo}, the learned self-supervised semantic maps and generated images based on the same layout in Fig.~\ref{fig:masks}, and more visual comparisons between our ablated models in Fig.~\ref{fig:abmasktwo}.

\subsection{Additional Results for Mask Clarity}
\label{app:more_clarity}
We present the full table of the quantitative results on mask clarity in Tab.~\ref{tab:sisapp}.
As for the visual comparison, we present the generated images from different semantic image synthesis and our \ac{ours} in Fig.~\ref{fig:t2iall}. It can be clearly seen that the self-supervised maps from our full model (\textbf{Ours}) produce images with recognizable objects and higher quality compared to the self-supervised maps from the \textbf{Base} (i.e., Tab.~\ref{tab:ablation}(a)). This again highlights the importance of the mask clarity. Moreover, it is evident that the semantic image synthesis methods highly rely on precisely labeled maps. When those are not available, their generation quality drops severely. This observation is also aligned with the quantitative results in Tab.~\ref{tab:sisapp}. On the other hand, our \ac{ours} still produces images with superior quality in this case due to the proposed \ac{ean} and \ac{sma} in Sec.~\ref{sec:method}.

\clearpage


\begin{table*}[!tbp]
\caption{The used hyperparameters for the proposed \ac{ours} in Sec. 4 experiments.}
\vspace{-6.2mm}
\begin{center}
    \begin{adjustbox}{max width=1.0\textwidth}
        \begin{tabular}{l c c c}
            \toprule
                \textbf{Dataset} & COCO-stuff 256$\times$256 & COCO-stuff 128$\times$128 & VG 256$\times$256 \\
                \textbf{Model} & \ac{ours}  &  \ac{ours}  &  \ac{ours}   \\
            \midrule
                \rowcolor{gray!20}Layout-Conditional Diffusion Model & & &  \\
                In Channels                &    3          &    3      &    3         \\
                Hidden Channels             &   256       &   128     &   256        \\
                Channel Multiply           & 1,1,2,2,4,4  & 1,1,2,3,4 & 1,1,2,2,4,4  \\
                Number of Residual Blocks  &    2     &    2     &    2         \\
                Dropout & 0 & 0 & 0 \\
                Diffusion Steps & 1000 & 1000 & 1000  \\
                Noise Schedule & linear & linear & linear  \\
                $\lambda$ & 0.001 & 0.001 & 0.001  \\
            \midrule
                \rowcolor{gray!20}Object Representation & & &  \\
                Class Embedding Dimension & 180 & 180 & 180  \\
                Style Embedding Dimension & 128 & 128 & 128  \\
                Maximum Number of Objects & 8 & 8 & 8  \\
                Maximum Number of Class Id & 184 & 184 & 179  \\
            \midrule
                \rowcolor{gray!20}\acl{ean} Module & & &  \\
                $\alpha$ & 0.5 & 0.5 & 0.5  \\
            \midrule
                \rowcolor{gray!20}\acl{sma} Module & & &  \\
                Attention Method & Styled-Mask  & Styled-Mask  & Styled-Mask   \\
                Number of Head Channels & 64 & 64 & 64 \\
            
            \midrule
                \rowcolor{gray!20}Training Hyperparameters & & &  \\
                Total Batch Size & 32 & 32 &  32  \\
                Number of GPUs & 4 & 4 & 4 \\
                Learning Rate & 1e-4 & 1e-4 & 1e-4 \\
                Mixed Precision Training   &   No   &   No   &   No     \\
                Weight Decay & 0 & 0 & 0  \\
                EMA Rate & 0.9999 & 0.9999 & 0.9999  \\
                Iterations & 1.25M & 600K & 1.45M  \\
            \midrule
                \rowcolor{gray!20}Sampling Hyperparameters & & &  \\
                Total Batch Size & 4 & 8 &  4  \\
                Number of GPUs & 1 & 1 & 1 \\
                Classifier-free Guidance $s$ & 1.5 & 1.5 & 1.0 \\
                Use DPM-Solver   &   True   &   True   &   True     \\
                DPM-Solver Algorithm & dpmsolver++ & dpmsolver++ & dpmsolver++  \\
                DPM-Solver Type & dpmsolver & dpmsolver & dpmsolver  \\
                DPM-Solver Skip Type & time\_uniform & time\_uniform & time\_uniform  \\
                DPM-Solver Step Method & singlestep & singlestep & singlestep  \\
                DPM-Solver ODE Order & 3 & 3 & 2  \\
                DPM-Solver Timesteps & 50 & 50 & 50  \\
            \bottomrule
        \end{tabular}
    \end{adjustbox}
\end{center}
\label{tab:trainhyper}
\end{table*}

\tabmetricsfcoco
\tabmetricsfvg
\tabsisapp

\clearpage
\figqualtwo
\figqualthree
\figvgqualone
\figvgqualtwo

\figstytwo
\figmasks
\figabmasktwo
\figttoiall

\end{document}